\documentclass[sigconf]{acmart}

\AtBeginDocument{%
  }

\copyrightyear{2022}
\acmYear{2022}
\setcopyright{acmcopyright}\acmConference[CIKM '22]{Proceedings of the 31st ACM International Conference on Information and Knowledge Management}{October 17--21, 2022}{Atlanta, GA, USA}
\acmBooktitle{Proceedings of the 31st ACM International Conference on Information and Knowledge Management (CIKM '22), October 17--21, 2022, Atlanta, GA, USA}
\acmPrice{15.00}
\acmDOI{10.1145/3511808.3557419}
\acmISBN{978-1-4503-9236-5/22/10}
\usepackage{graphicx,wrapfig}
\usepackage{subfigure}
\usepackage{enumitem}
\usepackage{pifont}
\usepackage{graphicx}
\usepackage{float}
\usepackage{multirow}
\usepackage[ruled]{algorithm2e}
\usepackage{color}
\definecolor{grey}{RGB}{211,211,211}
\newcommand{\modelname}{PASTEL}

\usepackage[normalem]{ulem}
\useunder{\uline}{\ul}{}

\newtheorem{myDef}{Definition}

\begin{document}

\title{Position-aware Structure Learning for Graph Topology-imbalance by Relieving Under-reaching and Over-squashing}

\author{Qingyun Sun}
\affiliation{%
  \institution{Beihang University}
  \city{Beijing}
  \country{China}
}
\email{sunqy@buaa.edu.cn}

\author{Jianxin Li}
\affiliation{%
  \institution{Beihang University}
  \city{Beijing}
  \country{China}
  }
\email{lijx@buaa.edu.cn}

\author{Haonan Yuan}
\affiliation{%
  \institution{Beihang University}
  \city{Beijing}
  \country{China}
}
\email{yuanhn@act.buaa.edu.cn}

\author{Xingcheng Fu}
\affiliation{%
 \institution{Beihang University}
  \city{Beijing}
  \country{China}
 }
\email{fuxc@act.buaa.edu.cn}

\author{Hao Peng}
\affiliation{%
  \institution{Beihang University}
  \city{Beijing}
  \country{China}
  }
\email{penghao@act.buaa.edu.cn}

\author{Cheng Ji}
\affiliation{%
  \institution{Beihang University}
  \city{Beijing}
  \country{China}
  \postcode{78229}
  }
\email{jicheng@act.buaa.edu.cn}

\author{Qian Li}
\affiliation{%
  \institution{Beihang University}
  \city{Beijing}
  \country{China}
  }
\email{liqian@act.buaa.edu.cn}

\author{Philip S. Yu}
\affiliation{%
  \institution{University of Illinois Chicago}
% \institution{Department of Computer Science, University of Illinois Chicago}
  \city{Chicago}
  \postcode{IL 60607}
  \country{USA}
  }
\email{psyu@uic.edu}

\renewcommand{\shortauthors}{Sun et al.}

\begin{abstract}
Topology-imbalance is a graph-specific imbalance problem caused by the uneven topology positions of labeled nodes, which significantly damages the performance of GNNs. 
What topology-imbalance means and how to measure its impact on graph learning remain under-explored. 
In this paper, we provide a new understanding of topology-imbalance from a global view of the supervision information distribution in terms of under-reaching and over-squashing, which motivates two quantitative metrics as measurements. 
In light of our analysis, we propose a novel position-aware graph structure learning framework named \modelname, which directly optimizes the information propagation path and solves the topology-imbalance issue in essence. 
Our key insight is to enhance the connectivity of nodes within the same class for more supervision information, thereby relieving the under-reaching and over-squashing phenomena. 
Specifically, we design an anchor-based position encoding mechanism, which better incorporates relative topology position and enhances the intra-class inductive bias by maximizing the label influence. 
% Specifically, we design an anchor-based position encoding mechanism, which better incorporates relative topology position information of nodes and enhances the intra-class inductive bias by maximizing the label influence. 
We further propose a class-wise conflict measure as the edge weights, which benefits the separation of different node classes. 
% Extensive experiments demonstrate the superiority of \modelname~in different data annotation scenarios. 
Extensive experiments demonstrate the superior potential and adaptability of \modelname~in enhancing GNNs' power in different data annotation scenarios. 
\end{abstract}


% \ccsdesc[500]{Mathematics of computing~Graph algorithms}

\maketitle

\section{Introduction}
\label{sec:introduction}
% Imbalance learning is a great importance for many types of data. 
% Graph and it suffers a specific imbalance problem, i.e., topology-imbalance. 
% Many data in real-world applications can be characterized with the graph structure from social networks~\cite{wasserman1994social} to proteins~\cite{jumper2021highly}, where nodes represent instances while edges indicate the relationship between instances. 
% Graph neural networks (GNNs)~\cite{wu2020comprehensive} have gained popularity over the past few years due to their versatility and success across a wide range of domains~\cite{nastase2015survey}. 
% Despite the impressive success of graph learning across the board research, 
Graph learning~\cite{zhang2020deep,li2022curvature,fu2021ace} has gained popularity over the past years due to its versatility and success in representing graph data across a wide range of domains~\cite{nastase2015survey,sun2020pairwise,gao2021room,yu2022cross,chen2022reinforced}. 
% The message-passing paradigm~\cite{gilmer2017neural} has been the ``battle horse'' of GNNs, which propagates the features on the graph by exchanging information between neighbor nodes. 
Graph Neural Networks (GNNs)~\cite{wu2020comprehensive,sun2021sugar} have been the ``battle horse'' of graph learning, which propagate the features on the graph by exchanging information between neighbors in a message-passing paradigm~\cite{gilmer2017neural}. 
Due to the asymmetric and uneven topology, learning on graphs by GNNs suffers a specific imbalance problem, i.e., topology-imbalance. 
Topology-imbalance~\cite{chen2021topology} is caused by the uneven position distribution of labeled nodes in the topology space, which is inevitable in real-world applications due to data availability and the labeling costs.  
% Topology-imbalance~\cite{chen2021topology} is caused by the uneven position distribution of labeled nodes with dense local areas in the topology space.  
% which is an inevitable problem in real-world applications due to data availability and the labeling costs. 
% expensive labeling costs. 
% Topology-imbalance is an inevitable problem in real-world applications due to data availability and the labeling costs. 
For example, we may only have information for a small group of users within a local community in social networks, resulting in a serious imbalance of labeled node positions. 
% it is difficult to construct a completely symmetric labeling set even with an abundant annotation budget. 
The uneven position distribution of labeled nodes leads to uneven information propagation, resulting in the poor quality of learned representations. 

% for regular data (e.g., texts and images)

% topology imbalance

% under-reaching 
% the over-squashing of information~\cite{alon2020bottleneck,topping2021understanding}
Although the imbalance learning on graphs has attracted many research interests in recent years, most of them focus on the class-imbalance issue~\cite{park2021graphens,wang2021distance}, i.e., the imbalanced number of labeled nodes of each class. 
The topology-imbalance issue is proposed recently and is still under-explored. 
The only existing work, ReNode~\cite{chen2021topology}, provides an understanding of the topology-imbalance issue from the perspective of label propagation and proposes a sample re-weighting method. 
% However, the performance and generalization ability of ReNode are limited by its strong homophily assumption. 
However, ReNode takes the node topological boundaries as decision boundaries based on a homophily assumption, which does not work with real-world graphs. 
The strong assumption leads to poor generalization and unsatisfied performance of ReNode (see Section~\ref{sec:cls_realworld}). 
% However, the strong homophily assumption of ReNode limits its generalization ability on real-world graphs, resulting in only slight performance improvement brought by ReNode (see Section~\ref{sec:cls_realworld}). 
There are \textbf{two remaining questions}: 
\textit{(1) Why does topology-imbalance affect the performance of graph representation learning?} and 
\textit{(2) What kind of graphs are susceptible to topology-imbalance?} 
To answer the above two questions, how to measure the influence of labeled nodes is the key challenge in handling topology-imbalance due to the complex graph connections and the unknown class labels for most nodes in the graph.

% GNNs are susceptible to a bottleneck when propagating messages originating from distant nodes and aggregating messages across a long path~\cite{alon2020bottleneck}. 

% GNNs are susceptible to two broad limitations~\cite{alon2020bottleneck,topping2021understanding}: \textit{under-reaching} (i.e., the depth of GNNs is insufficient to exchange information between distant nodes) and \textit{over-squashing} (i.e., certain edges act as the bottleneck for the information flow). 
\begin{figure}
    \centering
    \includegraphics[width=0.8\linewidth]{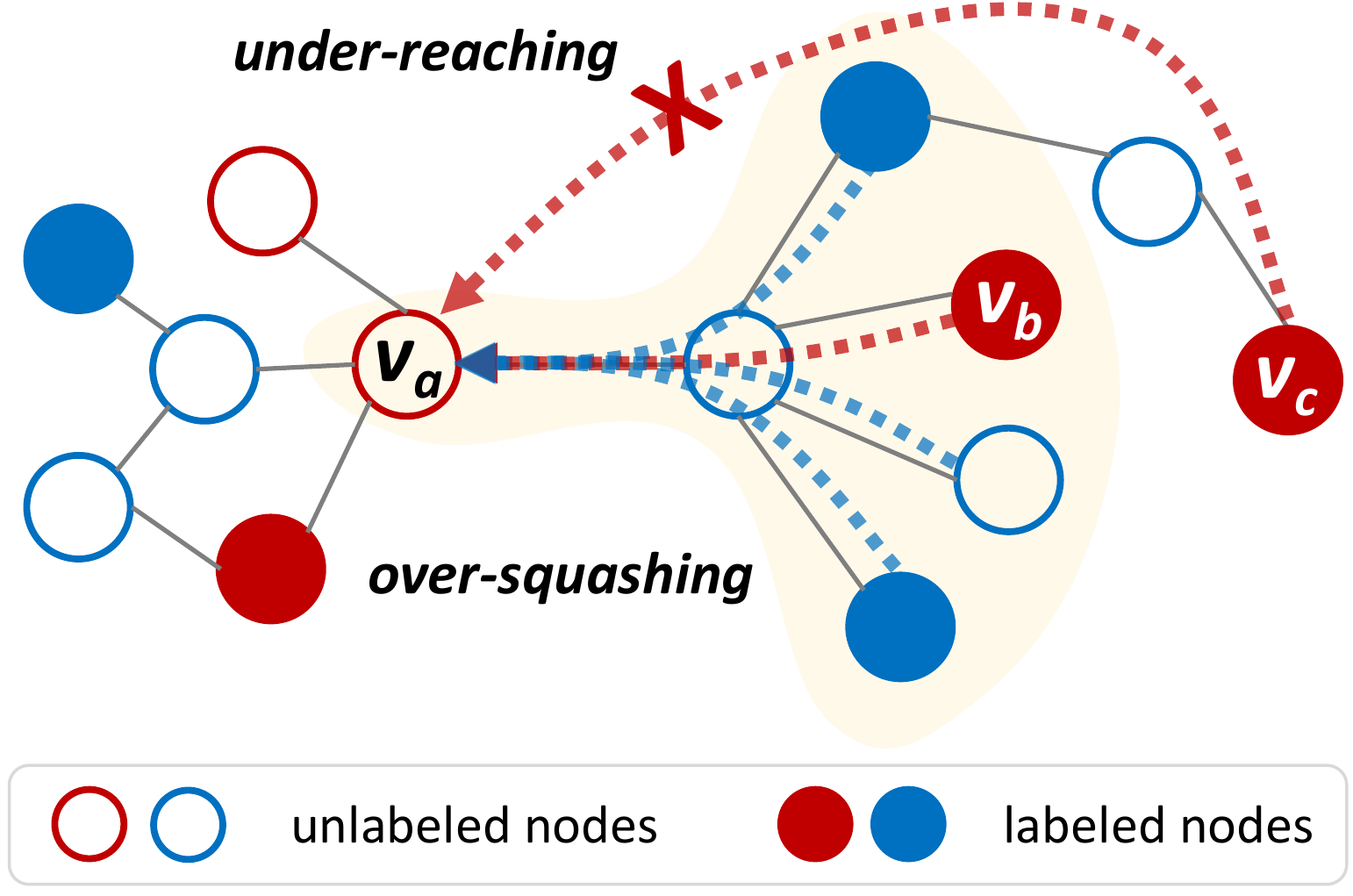}
    \caption{Schematic diagram of under-reaching and over-squashing in the topology-imbalance issue.}
    \label{fig:reach-squash}
\end{figure}
\textbf{New understanding for topology-imbalance. }
In this work, we provide a new understanding of the topology-imbalance issue from a global view of the supervision information distribution in terms of under-reaching and over-squashing: 
\textbf{(1) Under-reaching}: the influence of labeled nodes decays with the topology distance~\cite{buchnik2018bootstrapped}, resulting in the nodes far away from labeled nodes lack of supervision information. 
In Figure~\ref{fig:reach-squash}, the node $v_a$ cannot reach the valuable labeled node $v_c$ within the receptive field of the GNN model, resulting in the quantity of information it received is limited. 
\textbf{(2) Over-squashing}: the supervision information of valuable labeled nodes is squashed when passing across the narrow path together with other useless information. 
In Figure~\ref{fig:reach-squash}, the valuable supervision information of $v_b$ to $v_a$ is compressed into a vector together with the information of many nodes belonging to other classes, resulting in the quality of supervision information that $v_a$ received being poor. 
Then we introduce two metrics (reaching coefficient and squashing coefficient) to give a quantitative analysis of the relation between the learning performance, label positions, and graph structure properties. 
We further draw a conclusion that \textit{better reachability and lower squashing to labeled nodes lead to better classification performance for GNN models}. 

% Then we provide two quantitative metric for the under-reaching and over-squashing. 
\textbf{Present work. }
In light of the above analysis, we propose a \textbf{P}osition-\textbf{A}ware \textbf{ST}ructur\textbf{E} \textbf{L}earning method named \textbf{\modelname}, which directly optimizes the information propagation path and solves the problem of topology-imbalance issue in essence. 
The key insight of \modelname~is to enable nodes within the same class to connect more closely with each other for more supervision information. 
Specifically, we design a novel \textit{anchor-based position encoding mechanism} to capture the relative position between nodes and incorporate the position information into structure learning. 
Then we design a \textit{class-wise conflict measure} based on the Group PageRank, which measures the influence from labeled nodes of each class and acts as a guide to increase the intra-class connectivity via adjusting edge weight. 
The main contributions are as follows: 
\begin{itemize}[leftmargin=*]
    \item 
    We provide a new understanding of the topology-imbalance issue from the perspective of supervision information distribution in terms of under-reaching and over-squashing and provide two new quantitative metrics for them.  
    \item 
    % \modelname~properly incorporate position information of graphs into the model
    Equipped with the proposed position encodings and class-wise conflict measure, \modelname~can better model the relationships of node pairs and enhance the intra-class inductive bias by maximizing the label influence. 
        % Equipped with the proposed position encodings and class-wise conflict measure, \modelname~can better model the relationships of node pairs to balance the supervision information distribution in the topology space. 
    % \modelname~is model-agnostic and is scalable to large-scale graphs. 
    % It is sufficient to plug existing GNNs into the \modelname~framework to enhance their performances. 
    \item 
    % Extensive experiments on both synthetic and real-world graph dataset provide insightful analysis for the topology-imbalance problem. 
    Experimental results demonstrate that the proposed \modelname~enjoys superior effectiveness and indeed enhances the GNN model’s power for in-the-wild extrapolation. 

\end{itemize}
\section{Related Work}
\subsection{Imbalance Learning}
Imbalanced classification problems~\cite{sun2009classification,he2009learning} have attracted extensive research attention. 
Most existing works~\cite{haixiang2017learning,lin2017focal} focus on the class-imbalance problem, where the model performance is dominated by the majority class. 
% data synthesis
The class-imbalance learning methods can be roughly divided into two types: data-level re-sampling and algorithm-level re-weighting. 
\textbf{Re-sampling} methods re-sample~\cite{chawla2002smote,ando2017deep,xiaolong2019over} or augment data~\cite{park2021graphens} to balance the number of data for each class during the data selection phase. 
\textbf{Re-weighting} methods~\cite{cao2019learning,cui2019class,ren2018learning} adjust different weights to different data samples according to the number of data during the training phase. 

For the graph-specific topology-imbalance issue as mentioned in Section~\ref{sec:introduction}, directly applying these methods to the graph data fails to take the special topology properties into consideration. 
ReNode~\cite{chen2021topology} is the first work for the graph topology-imbalance issue, which follows the paradigm of classical re-weighting methods. 
Specifically, ReNode defines an influence conflict detection based metric and re-weights the labeled nodes based on their relative positions to class boundaries. 
However, ReNode is limited by its homophily assumption and only has a slight performance improvement.  
% on real graphs. 
% when the graph connectivity is poor. 
\textit{In this paper, \modelname~alleviates topology-imbalance by learning a new structure that maximizes the intra-class label influence, which can be seen as ``label re-distribution'' in the topology space. }

\subsection{Graph Structure Learning}
Graph structure learning~\cite{zhu2021deep} learns an optimized graph structure for representation learning and most of them aim to improve the robustness~\cite{jin2020graph,zheng2020robust} of GNN models. 
There are also some works~\cite{franceschi2019learning,chen2020iterative,topping2021understanding,chen2020label,sun2022graph} that utilize the structure learning to improve the graph representation quality. 
As for the over-squashing problem, \cite{wang2021combining} assigns different weights to edges connected to two nodes of the same class for better representations. 
However, \cite{wang2021combining} still fails with the issue of under-reaching. 
% (i.e., MPNNs fail to fully explore a graph when the depth is smaller than the diameter)~\cite{barcelo2020logical}. 
SDRF~\cite{topping2021understanding} rewires edges according to the Ricci curvatures to solve the over-squashing problem by only considering topology properties. 

Multiple measurements in existing structure learning works are leveraged for modeling node relations, including node features~\cite{zhao2021heterogeneous}, node degrees~\cite{jin2020graph}, node encodings~\cite{zhang2019hierarchical} and edge attributes~\cite{zheng2020robust}. 
The node positions play an important role in generating discriminative representations~\cite{you2019position} and are seldom considered in structure learning. 
\textit{In this work, we advance the structure learning strategy for the graph topology-imbalance issue and introduce a position-aware framework to better capture the nodes' underlying relations. }
\section{Understanding Topology-Imbalance}
% \section{Understanding Topology-Imbalance via Reaching and Squashing}
In this section, we provide a new understanding of the topology-imbalance issue in terms of under-reaching and over-squashing. 
% , and then give a quantitative analysis of the relations between them. 
Then we perform a quantitative analysis of the relations between them to answer two questions: 
% \begin{itemize}[leftmargin=*]
% \item \textbf{Q1:} Why does topology-imbalance affect the performance of graph representation learning? 
% \item \textbf{Q2:} What kind of graphs are susceptible to topology-imbalance?
% \end{itemize}

\noindent\textbf{Q1:} Why does topology-imbalance affect the performance of graph representation learning?  

\noindent\textbf{Q2:} What kind of graphs are susceptible to topology-imbalance?
% ``What kind of graphs are susceptible to the topology-imbalance issue?''

\subsection{Notations and Preliminaries}
Consider a graph $\mathcal{G}=\{\mathcal{V},\mathcal{E}\}$, where $\mathcal{V}$ is the set of $N$ nodes and $\mathcal{E}$ is the edge set.  
Let $\mathbf{A}\in \mathbb{R}^{N\times N}$ be the adjacency matrix and $\mathbf{X}\in \mathbb{R}^{N\times d_0}$ be the node attribute matrix, where $d_0$ denotes the dimension of node attributes. 
The diagonal degree matrix is denoted as $\mathbf{D}\in \mathbb{R}^{N\times N}$ where $\mathbf{D}_{ii}=\sum^{N}_{j=1}\mathbf{A}_{ij}$. 
% $Y=\{y_1,y_2,\cdots,y_m,y_{m+1},\cdots, y_N\}$ denotes the node label set. 
The graph diameter is denoted as $D_{\mathcal{G}}$. 
Given the labeled node set $\mathcal{V}_L$and their labels $\mathcal{Y}_L$ where each node $v_i$ is associated with a label $y_i$, \textit{semi-supervised node classification} aims to train a node classifier $f_{\theta}:v\rightarrow\mathbb{R}^{C}$ to predict the labels $\mathcal{Y}_U$ of remaining nodes $\mathcal{V}_U=\mathcal{V} \setminus \mathcal{V}_L$, where $C$ denotes the number of classes. 
we separate the labeled node set $\mathcal{V}_L$ into $\{\mathcal{V}^{1}_L, \mathcal{V}^2_L, \cdots, \mathcal{V}^{C}_L\}$, where $\mathcal{V}^i_L$ is the nodes of class $i$ in $\mathcal{V}_L$. 
%  containing $m$ nodes, 
% We denote the 

% \textbf{Graph Structure Learning.} 
% $f_{sl}:G\rightarrow G'$, 

% We can the graph structure learning into any existing GNN $f_{\theta}$ by $f_{\theta}:G\rightarrow\mathbb{R}^{C}$ as $f_{\theta}\circ f_{sl}:G\rightarrow\mathbb{R}^{C}$. 

\subsection{Understanding Topology-Imbalance via Under-reaching and Over-squashing} 
\label{sec:understand}
% \subsection{Quantitative Analysis for Topology-Imbalance.} 
% \begin{figure*}[htb]
% \centering
% \subfigure[Under-reaching and over-squashing.]{
% \begin{minipage}[t]{ 0.32\linewidth}
% \centering
% \includegraphics[width=\linewidth]{figs/reaching-squashing.pdf}
% \label{fig:reach_squash}
% \end{minipage}
% }
% \subfigure[Predictions of GCN with the same graph structure and different labeled nodes.]{
% \begin{minipage}[t]{ 0.32\linewidth}
% \centering
% \includegraphics[width=\linewidth]{figs/oneSBM-samestructure.pdf}
% \label{fig:samestructure}
% \end{minipage}
% }
% \subfigure[Predictions of GCN with and the same labeled nodes and different graph structures.]{
% \begin{minipage}[t]{ 0.32\linewidth}
% \centering
% \includegraphics[width=\linewidth]{figs/oneSBM-samelabels.pdf}
% \label{fig:samelabels}
% \end{minipage}
% }
% \caption{
% Topology-imbalance affects GNNs' performance by under-reaching and over-squashing. 
% (a) Schematic diagram of under-reaching and over-squashing in the topology-imbalance issue. 
% (b) The relation between the reaching coefficient $RC$, the squashing coefficient $SC$ and GCN's performance with the same graph structure and different labeled nodes. 
% (c) The relation between the reaching coefficient $RC$, the squashing coefficient $SC$ and GCN's performance with different graph structures and the same labeled nodes. 
% }
% \end{figure*}

% In this subsection, we provide a new understanding of the topology-imbalance issue from a global view of the supervision information distribution in terms of under-reaching and over-squashing. 

In GNNs, node representations are learned by aggregating information from valuable neighbors. 
The quantity and quality of the information received by the nodes decide the expressiveness of their representations. 
% Intuitively, the information quantity can be seen as the reachability between nodes and their, and the information quality can be seen as the useful information left over after squashing along the graph topology. 
% Intuitively, 
We perceive the imbalance of the labeled node positions affects the performance of GNNs for two reasons:

\noindent(1) \textbf{Under-reaching}: 
The influence from labeled nodes decays with the topology distance~\cite{buchnik2018bootstrapped}, resulting in that the nodes far away from labeled nodes lack supervision information. 
When the node can't reach enough valuable labeled nodes within the receptive field of the model, the quantity of information it received is limited. 
% As shown in Figure~\ref{fig:reach_squash}, when the node $v_i$ can't reach enough valuable labeled nodes (e.g., $v_a$ and $v_b$) within a number of hops, the quantity of information that nodes received is limited. 

\noindent(2) \textbf{Over-squashing}: 
The receptive field of GNNs is exponentially-growing and all information is compressed into fixed-length vectors~\cite{alon2020bottleneck}. 
The supervision information of valuable labeled nodes is squashed when passing across the narrow path together with other useless information. 
% As shown in Figure~\ref{fig:reach_squash}, the quality of information that $v_i$ received is poor. 
% ReNode~\cite{chen2021topology} provides a influence conflict based topology-imbalance metric named Totoro, which is based on the graph homophily assumption and not applicable to all graphs. 

\subsection{Quantitative Analysis}
\label{sec:quantitative}

To provide quantitative analysis for topology-imbalance, we propose two metrics for reachability and squashing. 
First, we define a reaching coefficient based on the shortest path, which determines the minimum layers of GNNs to obtain supervision information: 

\begin{myDef}[\textbf{Reaching coefficient}]
Given a graph $\mathcal{G}$ and labeled node set $\mathcal{V}_L$, the reaching coefficient $RC$ of $\mathcal{G}$ is the mean length of the shortest path from unlabeled nodes to the labeled nodes of their corresponding classes: 
\begin{equation}
    RC=\frac{1}{|\mathcal{V}_U|}\sum_{v_i\in \mathcal{V}_U}\frac{1}{|\mathcal{V}^{y_i}_L|}\sum_{v_j\in \mathcal{V}^{y_i}_L}\left (1-\frac{\log |\mathcal{P}_{sp}(v_i,v_j)|}{\log D_{\mathcal{G}}}\right ),
\end{equation}
where $\mathcal{V}^{y_i}_L$ denotes the nodes in $\mathcal{V}_L$ whose label is $y_i$, $\mathcal{P}_{sp}(v_i,v_j)$ denotes the shortest path between $v_i$ and $v_j$, and $|\mathcal{P}_{sp}(v_i,v_j)|$ denotes its length, and $D_{\mathcal{G}}$ is the diameter of graph $\mathcal{G}$. 
Specifically, for the unconnected $v_i$ and $v_j$, we set the length of their shortest path as $D_{\mathcal{G}}$. 
\end{myDef}
The reaching coefficient reflects how long the the distance when the GNNs passes the valuable information to the unlabeled nodes. 
Note that $RC\in [0,1)$ and larger $RC$ means better reachability. 

For the quantitative metric of over-squashing, we define a squashing coefficient using the Ricci curvature to formulate it from a geometric perspective. 
The Ricci curvature~\cite{ollivier2009ricci} reflects the change of topology properties of the two endpoints of an edge, where the negative $Ric(v_i,v_j)$ means that the edge behaves locally as a shortcut or bridge and positive $Ric(v_i,v_j)$ indicates that locally there are more triangles in the neighborhood of $v_i$ and $v_j$~\cite{ni2015ricci,topping2021understanding}. 

\begin{myDef}[\textbf{Squashing coefficient}]
Given a graph $\mathcal{G}$, the squashing coefficient $SC$ of $\mathcal{G}$ is the mean Ricci curvature of edges on the shortest path from unlabeled nodes to the labeled nodes of their corresponding classes: 
% is the mean Ricci curvature of all edges:
\begin{equation}
    SC=\frac{1}{|\mathcal{V}_U|}\sum_{v_i\in \mathcal{V}_U}\frac{1}{|\mathcal{N}_{y_i}(v_i)|}\sum_{v_j\in \mathcal{N}_{y_i}(v_i)}\frac{\sum_{e_{kt}\in \mathcal{P}_{sp}(v_i,v_j)} Ric(v_k, v_t)}{|\mathcal{P}_{sp}(v_i,v_j)|},
\end{equation}
% \begin{equation}
%     SC=\frac{1}{|\mathcal{E}|}\sum_{e_{i,j}\in \mathcal{E}}Ric(i,j),
% \end{equation}
where $\mathcal{N}_{y_i}(v_i)$ denotes the labeled nodes of class $y_i$ that can reach $v_i$, $Ric(\cdot,\cdot)$ denotes the Ricci curvature, and $|\mathcal{P}_{sp}(v_i,v_j)|$ denotes the length of shortest path between $v_i$ and $v_j$. 
\end{myDef}
We leverage the \textit{Ollivier-Ricci curvature}~\cite{ollivier2009ricci} as $Ric(\cdot,\cdot)$ here: 
% which can be calculated by: 
\begin{equation}
    Ric(v_k,v_t)=\frac{Wasserstein(mass_k, mass_t)}{d_{geo}(v_k,v_t)}, 
\end{equation}
where $Wasserstein(\cdot,\cdot)$ is the Wasserstein distance, $d_{geo}(\cdot,\cdot)$ is the geodesic distance function, and $mass_k$ is the mass distribution~\cite{ollivier2009ricci} of node $v_k$. 
% The squashing coefficient reflects the \todo{}. 
Note that $SC$ can be either positive or negative and larger $SC$ means lower squashing because the ring structures are more friendly for information sharing.  

% Considering the 
\begin{figure}[!htp]
    \centering
    \includegraphics[width=0.95\linewidth]{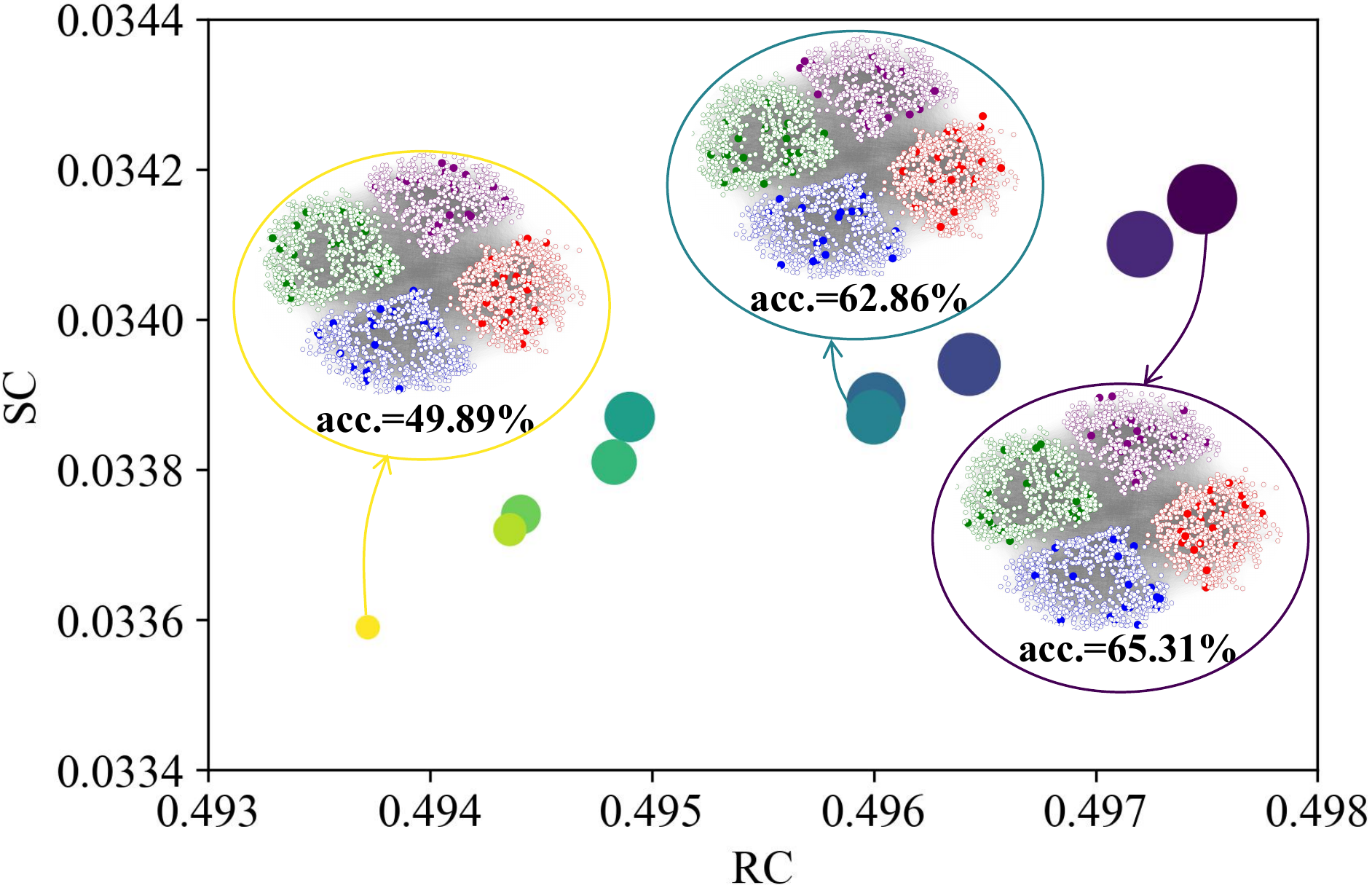}
    \caption{Predictions of GCN with the same graph structure and different labeled nodes.}
    \label{fig:samestructure}
\end{figure}

\begin{figure}[t]
    \centering
    \includegraphics[width=0.95\linewidth]{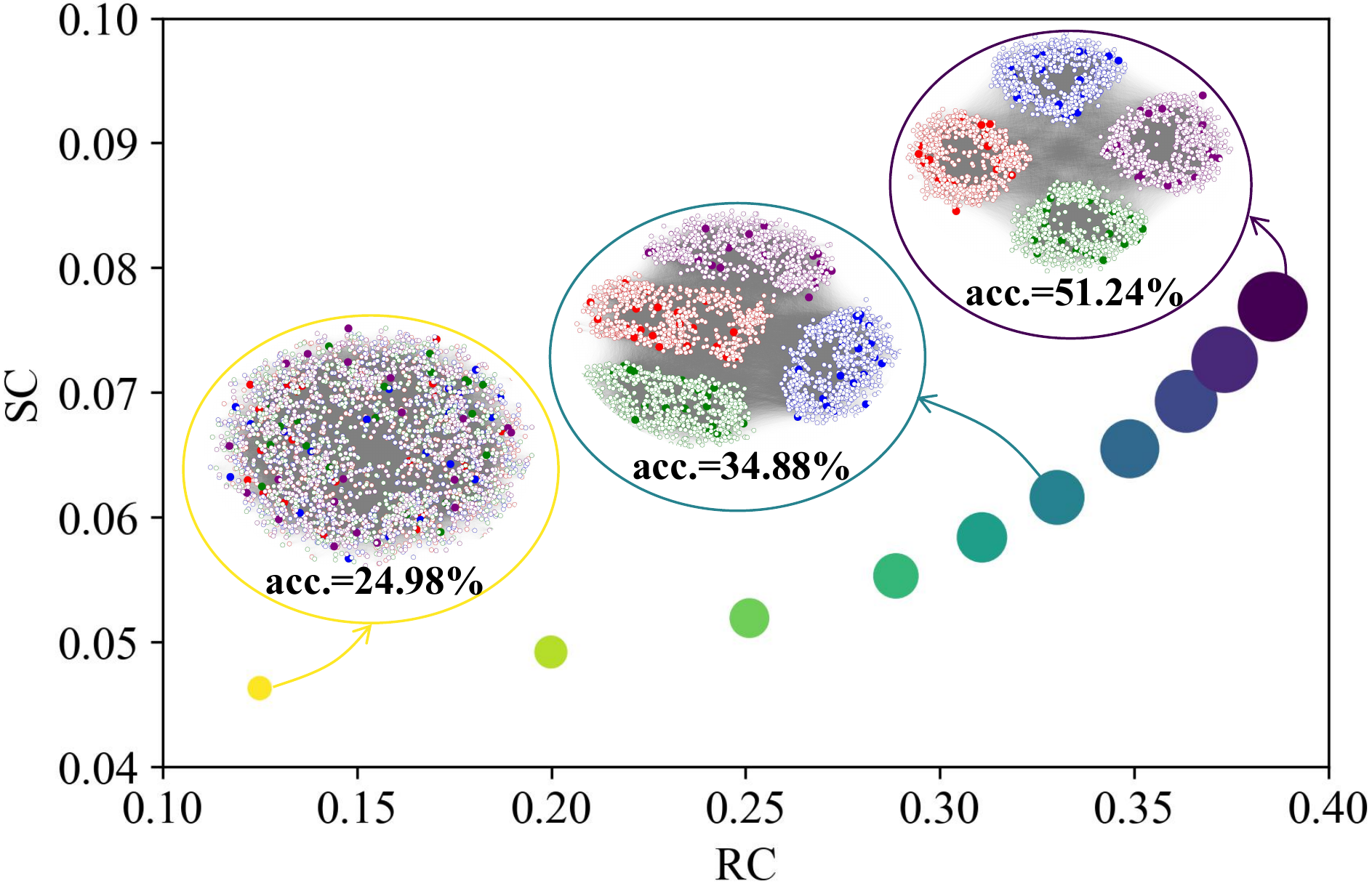}
    \caption{
    Predictions of GCN with and the same labeled nodes and different graph structures. 
    }
    \label{fig:samelabels}
\end{figure}

In Figure~\ref{fig:samestructure} and Figure~\ref{fig:samelabels}, we show the relation between the reaching coefficient $RC$, the squashing coefficient $SC$, and the classification accuracy. 
The higher the accuracy, the darker and larger the corresponding scatter. 
% where the node size and the shade of color represent the magnitude of classification accuracy. 
First, we analyze the performance of GCN when trained on the same graph structure but with different labeled nodes. 
In Figure~\ref{fig:samestructure}, we generate a synthetic graph by the Stochastic Block Model (SBM)~\cite{holland1983stochastic} with 4 classes and 3,000 nodes. 
We randomly sample some nodes as the labeled nodes 10 times and scatter the classification accuracy in Figure~\ref{fig:samestructure}. 
We can observe that even for the same graph structure, the difference in positions of labeled nodes may bring up to 15.42\% difference in accuracy. 
There is a significant positive correlation between the reaching coefficient, the squashing coefficient, and the model performance. 

%  and better reachability (smaller $RC$) leads better performance. 

Then we analyze the performance of GCN when trained with the same labeled nodes but on different graph structures. 
In Figure~\ref{fig:samelabels}, we set the labeled nodes to be the same and generate different structures between them by controlling the edge probability between communities in the SBM model. 
We can observe that with the same supervision information, there is up to a 26.26\% difference in accuracy because of the difference in graph structures. 
There is also a significant positive correlation between the reaching coefficient, the squashing coefficient, and the model performance. 
When the graph shows better community structure among nodes of the same class, the node representations can be learned better. 

% and lower squashing (larger $SC$) leads better performance. 

% \begin{figure}
%     \centering
%     \includegraphics[width=0.9\linewidth]{figs/sp_GNNacc.pdf}
%     \caption{Caption}
%     \label{fig:sp_GNNacc}
% \end{figure}

% Therefore, we make the following \todo{observations / assumptions / conclusions}. 
% \begin{myAssumption}
% Better reachability to labeled nodes leads to better classification performance for GNN models. 
% \end{myAssumption}
% \begin{myAssumption}
% Lower squashing of graph data leads to better classification performance for GNN models. 
% \end{myAssumption}

\textbf{Therefore, we make the following conclusions: }
% (1) The position of the labeled node matters for the performance of representation learning. 
% (1) The proposed two metrics can effectively reflect the degree of topology-imbalance and are also the reason for the effect caused by topology-imbalance. (Q1)
(1)
% Topology-imbalance is closely related with under-reaching and over-squashing. 
% Understanding topology-imbalance from under-reaching and over-squashing is reasonable. 
% Under-reaching and over-squashing are the reasons for the effect caused by topology-imbalance. 
Topology-imbalance hurts the performance of graph learning in the way of under-reaching and over-squashing. 
(for Q1) 
(2) The proposed two quantitative metrics can effectively reflect the degree of topology-imbalance. 
Graph with poor reachability (i.e., smaller $RC$) and stronger squashing (i.e., smaller $SC$) is more susceptible to topology-imbalance. (for Q2)
(3) Optimizing the graph structure can effectively solve the topology-imbalance issue. 
% (3) Graph structure with better reachability (i.e., larger $RC$) and lower squashing (i.e., larger $SC$) to labeled nodes leads to better classification performance for GNN models. 
% The quantitative analysis of the relation between reachability, squashing and classification accuracy provides the guideline for designing the framework of \modelname. 
The above conclusions provide the guideline for designing the framework of \modelname, i.e., balance the supervision information distribution by learning a structure with better reachability and lower squashing.

\section{Alleviate Topology-Imbalance by Structure Learning}

\begin{figure*}[t]
    \centering
    \includegraphics[width=\linewidth]{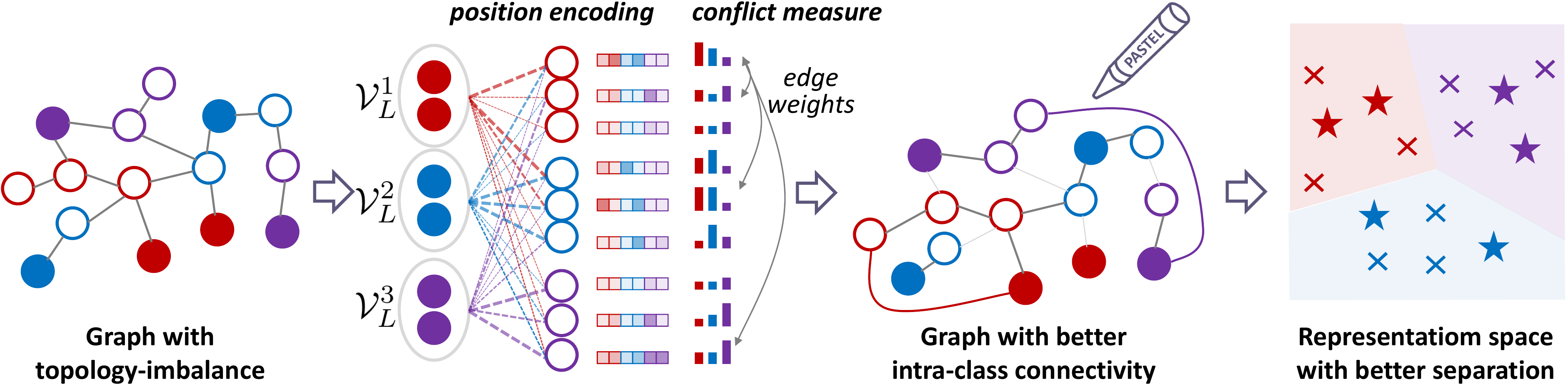}
    \caption{
    Overall architecture of \modelname. 
    \modelname~encodes the relative position between nodes with the labeled nodes as anchor sets \{$\mathcal{S}$\} and incorporates the position information with node features for structure learning. 
    For each pair of nodes, \modelname~uses the class-wise conflict measure as the edge weights to learn a graph with better intra-class connectivity. 
    }
    \label{fig:framework}
\end{figure*}

In this section, we introduce \textbf{\modelname}, a \textbf{P}osition-\textbf{A}ware \textbf{ST}ructur\textbf{E} \textbf{L}earning framework, to optimize the information propagation path directly and address the topology-imbalance issue in essence. 
% for general GNN models in a plug-and-play manner. 
In light of the analysis in Section~\ref{sec:understand}, \modelname~aims to learn a better structure that increases the intra-class label influence for each class and thus relieves the under-reaching and over-squashing phenomena. 
The overall architecture of \modelname~is shown in Figure~\ref{fig:framework}.

\subsection{Position-aware Structure Learning}
\label{sec:position_encoding}
% We learn the graph structure in a metric learning way and train a probability function for connecting two nodes. 
% To form structure with better intra-class connectivity, we use an anchor-based position encoding method to capture the relative position of a given node with respect to all the labeled nodes of the graph. 
To form structure with better intra-class connectivity, we use an anchor-based position encoding method to capture the topology distance between unlabeled nodes to labeled nodes. 
Then we incorporate both the merits of feature information as well as topology information to learn the refined structure. 
% inductive. 

% Holistic Position Inference

% in the node feature view. 

% by incorporating both the merits of global information as well as local information. 

% \textbf{relative position encoding}
% encode the relative distance of any two positions. 

\textbf{Anchor-based Position Encoding. }
Inspired by the position in transformer~\cite{vaswani2017attention,shaw2018self}, we use an anchor-based position encoding method to capture the relative position of unlabeled nodes with respect to all the labeled nodes of the graph. 
Since we focus on maximizing the reachability between unlabeled nodes and labeled nodes within the same class, we directly separate the labeled node set $\mathcal{V}_L$ into $C$ anchor sets $\{\mathcal{V}^{1}_L, \mathcal{V}^2_L, \cdots, \mathcal{V}^{C}_L\}$, where each subset $\mathcal{V}^c_L$ denotes the labeled nodes whose labels are $c$. 
The class-wise anchor sets help distinguish the information from different classes rather than treating all the anchor nodes the same and ignoring the class difference as in~\cite{you2019position}. 
Concretely, for any node $v_i$, we consider a function $\phi(\cdot, \cdot)$ which measures the position relations between $v_i$ and the anchor sets in graph $\mathcal{G}$. 
The function can be defined by the connectivity between the nodes in the graph. 
% In this paper, we choose $\phi(\cdot, \cdot)$ to be the mean distance of the shortest path between $v_i$ and anchor sets: 
\begin{equation}
    \mathbf{p}_i=\left(\phi\left(v_i, \mathcal{V}^1_L\right),\phi\left(v_i, \mathcal{V}^2_L\right),\cdots,\phi\left(v_i, \mathcal{V}^C_L\right)\right),
\end{equation}
where $\phi(v_i, \mathcal{V}^c_L)$ is the position encoding function defined by the connectivity between the node $v_i$ and the anchor set $\mathcal{V}^c_L$ in graph. 
Here we choose $\phi(v_i, \mathcal{V}^c_L)$ to be the mean length of shortest path between $v_i$ and nodes in $\mathcal{V}^c_L$ if two nodes are connected: 
\begin{equation}
    \phi(v_i, \mathcal{V}^c_L)=\frac{\sum_{v_j \in \mathcal{N}_{c}(v_i)}|\mathcal{P}_{sp}(v_i,v_j)|}{|\mathcal{N}_{c}(v_i)|},
\end{equation}
where $\mathcal{N}_{c}(v_i)$ is the nodes connected with $v_i$ in $\mathcal{V}^c_L$ and $|\mathcal{P}_{sp}(v_i,v_j)|$ is the length of shortest path between $v_i$ and $v_j$. 
Then we transform the position encoding into the $d_0$ dimensional space: 
\begin{equation}
\label{eq:position_encoding}
    \mathbf{h}^p_i=\mathbf{W}_{\phi}\cdot\mathbf{p}_i, 
\end{equation}
where $\mathbf{W}_{\phi}$ is a trainable vector. 
If two nodes have similar shortest paths to the anchor sets, their position encodings are similar. 
\textbf{Position-aware Metric Learning. }
After obtaining the position encoding, we use a metric function that accounts for both node feature information and the position-based similarities to measure the possibility of edge existence. 
\modelname~is agnostic to various similarity metric functions and we choose the widely used multi-head cosine similarity function here: 
\begin{equation}
\label{eq:metric}
    a^P_{ij}=\frac{1}{m}\sum^{m}_{h=1}\cos\left(\mathbf{W}_h \cdot \left(\mathbf{z}_i||\mathbf{h}^p_i\right), \mathbf{W}_h\cdot \left(\mathbf{z}_j||\mathbf{h}^p_j\right)\right),
\end{equation}
where $m$ is the number of heads, $\mathbf{W}_h$ is the weight matrix of the $h$-th head, $\mathbf{z}_i$ denotes the representation vector of node $v_i$ and $||$ denotes concatenation. 
The effectiveness of the position-aware structure learning is evaluated in Section~\ref{sec:abla_position}. 

% \textbf{absolute position encoding}
% give each position an embedding\

% To speed up the computation, we hence proceed to extract a symmetric sparse non-negative adjacency matrix A from S by masking off (i.e., set to zero) those elements in $\mathbf{A}^*$ which are smaller than a predefined non-negative threshold $a_0$. 

% \subsection{Distance-wise conflict measure}
\subsection{Class-wise Conflict Measure}
\label{sec:conflict}
We aim to increase the intra-class connectivity among nodes, thereby increasing the supervision information they received and their influence on each other. 
Here we propose a class-wise conflict measure to guide what nodes should be more closely connected. 
According to the inherent relation of GNNs with Label Propagation~\cite{wang2021combining,chen2021topology}, we use the \textit{Group PageRank}~\cite{chen2020distance} as a conflict measure between nodes. 
Group PageRank (GPR) extends the traditional PageRank\cite{page1999pagerank} into a label-aware version to measure the supervision information from labeled nodes of each class. 
Specifically, for class $c\in\{1,2,\cdots, C\}$, the corresponding GPR matrix is
\begin{equation}
    \mathbf{P}^{gpr}(c) = (1-\alpha)\mathbf{A}'\mathbf{P}^{gpr}(c) + \alpha \mathbf{I}_c,
\end{equation}
where $\mathbf{A}'=\mathbf{A}\mathbf{D}^{-1}$, 
% $\mathbf{D}$ is the degree matrix, 
$\alpha$ is the random walk restart probability at a random node in the group and $\mathbf{I}_c\in \mathbb{R}^{n}$ is the teleport vector:
\begin{equation}
    \mathbf{I}^i_c = \left\{\begin{matrix}
 \frac{1}{|\mathcal{V}^c_L|}, & if~y_i=c \\
  0,& otherwise
\end{matrix}\right.
\end{equation}
where $|\mathcal{V}^c_L|$ is the number of labeled nodes with class $c$. 
We calculate the GPR for each group individually and then concatenate all the GPR vectors to form a final GPR matrix $\mathbf{P}^{gpr}\in \mathbb{R}^{N\times C}$ as in~\cite{chen2020distance}: 
\begin{equation}
\label{eq:gpr_matrix}
    \mathbf{P}^{gpr}=\alpha\left(\mathbf{E}-\left(1-\alpha\right)\mathbf{A}'\right)^{-1}\mathbf{I}^*,
\end{equation}
where $\mathbf{E}$ is the unit matrix of nodes and $\mathbf{I}^*$ is the concatenation of $\{\mathbf{I}_c, c=1,2,\cdots,C\}$. 
Under $\mathbf{P}^{gpr}$, node $v_i$ corresponds to a GPR vector $\mathbf{P}^{gpr}_{i}$ (the $i$-th row of $\mathbf{P}^{gpr}$), where the $c$-th dimension represents the the supervision influence of labeled nodes of class $c$ on node $v_i$. 
The GPR value contains not only the global topology information but also the annotation information. 
% of labeled nodes. 

For each node pair nodes $v_i$ and $v_j$, we use the Kullback Leiber (KL) divergence of their GPR vectors to measure their conflict when forming an edge: 
\begin{equation}
    \kappa_{ij}=\operatorname{KL}\left(\mathbf{P}^{gpr}_{i}, \mathbf{P}^{gpr}_{j}\right). 
\end{equation}
The distance of GPR vectors reflects the influence conflict of different classes when exchanging information. 
We use a cosine annealing mechanism to calculate the edge weights by the relative ranking of the conflict measure: 
\begin{equation}
\label{eq:conflict}
    w_{ij}=\frac{1}{2}\left [-\cos{\frac{\operatorname{Rank}(\kappa_{ij})}{|\mathcal{V}|\times |\mathcal{V}|}*\pi}+1\right ], 
\end{equation}
where $Rank(\cdot)$ is the ranking function according to the magnitude. 
The more conflicting the edge is, the less weight is assigned to it. 
With the class-wise conflict measure, we aim to learn a graph structure that makes the GPR vectors of nodes have ``sharp'' distributions focusing on their ground-truth classes. 
Then $w_{ij}$ is used as the connection strength of edge $e_{ij}$, with the corresponding element $\tilde{a}^P_{ij}$ in the adjacency matrix being: 
\begin{equation}
\label{eq:conflict_weight}
    \tilde{a}^P_{ij}=w_{ij}\cdot a^P_{ij}.
\end{equation}
The effectiveness of the class-wise conflict measure is evaluated in Section~\ref{sec:abla_conflict} and the change of GPR vectors is shown in Section~\ref{sec:gpr_visual}. 

\begin{algorithm}[!t]
\caption{The overall process of \modelname}
\label{alg:training}
\LinesNumbered
\KwIn{Graph $\mathcal{G}=\{\mathcal{V},\mathcal{E}\}$ with node labels $\mathcal{Y}$; Number of heads $m$; Number of training epochs $E$; Structure fusing coefficients $\lambda_1, \lambda_2$; Loss coefficients $\beta_1, \beta_2, \beta_3$}
\KwOut{Optimized graph $\mathcal{G}^*=(\mathbf{A}^*, \mathbf{X})$, predicted label $\hat{\mathcal{Y}}$}
% Let $\phi \leftarrow \phi_0$, $\theta \leftarrow \theta_0$\\
Parameter initialization;\\
% \tcp{Train \modelname}
\For{$e=1,2,\cdots,E$}{
\tcp{Learn position-aware graph structure}
Learn position encodings $\mathbf{h}^p_i$ $\gets$ Eq.~\eqref{eq:position_encoding};\\
% \tcp{Position-aware metric learning}
Learn edge possibility $a^P_{ij}$ $\gets$ Eq.~\eqref{eq:metric};\\
% \tcp{Measure class-wise conflict}
Calculate the Group PageRank matrix $\mathbf{P}^{gpr}\gets$ Eq.~\eqref{eq:gpr_matrix};\\
Calculate the class-wise conflict measure $w_{ij}\gets$ Eq.~\eqref{eq:conflict};\\
Obtain position-aware structure $\mathbf{A}_P$ $\gets$  Eq.~\eqref{eq:positionA};\\
% \tcp{Obtain optimized structure} 
\tcp{Learn node representations}
Obtain the optimized structure $\mathbf{A}^*$ $\gets$ Eq.~\eqref{eq:A*};\\
Calculate representations and labels $\mathbf{Z}$, $\hat{\mathcal{Y}}$ $\gets$ Eq.~\eqref{eq:representation_predict};\\
% Predict node labels $\hat{\mathcal{Y}}$ $\gets $ Eq.~\eqref{eq:predict};\\
\tcp{Optimize}
Calculate the losses $\mathcal{L}_{cls}\gets$ Eq.~\eqref{eq:loss_cls}, $\mathcal{L}_{smooth}\gets$Eq.~\eqref{eq:loss_smooth}, $\mathcal{L}_{con}\gets$Eq.~\eqref{eq:loss_connectivity}, and  $\mathcal{L}_{spar}\gets$Eq.~\eqref{eq:loss_sparsity}; \\
% Calculate the classification loss and graph regularization losses $\mathcal{L}_{cls}\gets$ Eq.~\eqref{eq:loss_cls}, $\mathcal{L}_{smooth}\gets$Eq.~\eqref{eq:loss_smooth}, $\mathcal{L}_{con}\gets$Eq.~\eqref{eq:loss_connectivity}, and  $\mathcal{L}_{spar}\gets$Eq.~\eqref{eq:loss_sparsity}; \\
Update model parameters to minimize $\mathcal{L}$ $\gets$ Eq.~\eqref{eq:loss}. 
}
\end{algorithm}

\subsection{Learning with the Optimized Structure}
With the above structure learning strategy, we can obtain a position-aware adjacency $\mathbf{A}_{P}$ with maximum intra-class connectivities: 
\begin{equation}
\label{eq:positionA}
    \mathbf{A}_{P}=\{\tilde{a}^P_{ij}, i,j\in \{1,2,\cdots,N\}\}. 
\end{equation}
% After learned the position-aware structure , we utilize it for node representation learning. 
The input graph structure determines the learning performance to a certain extent. 
Since the structure learned at the beginning is of poor quality, directly using it may lead to non-convergence or unstable training of the whole framework. 
We hence incorporate the original graph structure $\mathbf{A}$ and a structure in a node feature view $\mathbf{A}_N$ as supplementary to formulate an optimized graph structure $\mathbf{A}^*$. 
Specifically, we also learn a graph structure $\mathbf{A}_{N}=\{a^N_{ij}, i,j\in \{1,2,\cdots,N\}\}$ in a node feature view with each element being:  
\begin{equation}
    a^{N}_{ij}=\frac{1}{m}\sum^{m}_{h=1}\cos\left(\mathbf{W}_h \cdot \left(\mathbf{x}_i||\mathbf{h}^{p_0}_i\right), \mathbf{W}_h\cdot \left(\mathbf{x}_j||\mathbf{h}^{p_0}_j\right)\right), 
\end{equation}
where $\mathbf{x}_i$ is the feature vector of node $v_i$ and $\mathbf{h}^{p_0}_i$ is the position encoding with the original structure. 
% We incorporate the original graph structure as a supplementary to 
Then we can formulate an optimized graph structure $\mathbf{A}^*$ with respect to the downstream task: 
\begin{equation}
\label{eq:A*}
    \mathbf{A}^*=\lambda_1\mathbf{D}^{-\frac{1}{2}}\mathbf{A}\mathbf{D}^{-\frac{1}{2}}+\left(1-\lambda_1\right)\left(\lambda_2 f\left(\mathbf{A}_{N}\right)+\left(1-\lambda_2\right)f\left(\mathbf{A}_P\right)\right),  
\end{equation}
where $f(\cdot)$ denotes the row-wise normalization function, $\lambda_1$ and $\lambda_2$ are two constants that control the contributions of original structure and feature view structure, respectively. 
Here we use a dynamic decay mechanism for $\lambda_1$ and $\lambda_2$ to enable the position-aware structure $\mathbf{A}_P$ to play a more and more important role during training. 

To control the quality of learned graph structure, we impose additional constraints on it following~\cite{kalofolias2016learn,chen2020iterative} in terms of smoothness, connectivity, and sparsity: 
% \begin{equation}
% \label{eq:loss_smooth}
%     \mathcal{L}_{smooth}=\frac{1}{N^2}{\rm tr}\left(\mathbf{X}^{T}\mathbf{L}^*\mathbf{X}\right),
% \end{equation}
% \begin{equation}
% \label{eq:loss_connectivity}
%     \mathcal{L}_{con}=\frac{1}{N}\mathbf{1}^{T}\log(\mathbf{A}^*\mathbf{1}),
% \end{equation}
% \begin{equation}
% \label{eq:loss_sparsity}
%     \mathcal{L}_{spar}=\frac{1}{N^2}||\mathbf{A}^*||^{2}_{F},
% \end{equation}
\begin{align}
\label{eq:loss_smooth}
    \mathcal{L}_{smooth}&=\frac{1}{N^2}{\rm tr}\left(\mathbf{X}^{T}\mathbf{L}^*\mathbf{X}\right),\\
    \label{eq:loss_connectivity}
    \mathcal{L}_{con}&=\frac{1}{N}\mathbf{1}^{T}\log(\mathbf{A}^*\mathbf{1}),\\
    \label{eq:loss_sparsity}
    \mathcal{L}_{spar}&=\frac{1}{N^2}||\mathbf{A}^*||^{2}_{F},
\end{align}
where $\mathbf{L}^*=\mathbf{D}^*-\mathbf{A}^*$ is the Laplacian of $\mathbf{A}^*$ and $\mathbf{D}^*$ is the degree matrix of $\mathbf{A}^*$. 
To speed up the computation, we extract a symmetric sparse non-negative adjacency matrix by masking off (i.e., set to zero) those elements in $\mathbf{A}^*$ which are smaller than a predefined non-negative threshold $a_0$. 
Then $\mathcal{G}^*=(\mathbf{A}^*, \mathbf{X})$ is input into the $\operatorname{GNN-Encoder}$ for the node representations $\mathbf{Z}\in\mathbb{R}^{N\times d}$, predicted labels $\hat{y}$ and classification loss $\mathcal{L}_{cls}$: 
% \begin{equation}
% \label{eq:node_representations}
%     \mathbf{Z}=\operatorname{GNN-Encoder}(\mathbf{A}^*, \mathbf{X}),
% \end{equation}
% \begin{equation}
% \label{eq:predict}
%     \hat{\mathcal{Y}}=\operatorname{Classifier}(\mathbf{Z}),
% \end{equation}
% \begin{equation}
% \label{eq:loss_cls}
%     \mathcal{L}_{cls}=\operatorname{Cross-Entropy}(\mathcal{Y}, \hat{\mathcal{Y}}).
% \end{equation}
\begin{equation}
 \label{eq:representation_predict}
    \mathbf{Z}=\operatorname{GNN-Encoder}(\mathbf{A}^*, \mathbf{X}),\hat{\mathcal{Y}}=\operatorname{Classifier}(\mathbf{Z}),
\end{equation}
\begin{equation}
    \label{eq:loss_cls}
    \mathcal{L}_{cls}=\operatorname{Cross-Entropy}(\mathcal{Y}, \hat{\mathcal{Y}}).
\end{equation}
The overall loss is defined as the combination of the node classification loss and graph regularization loss: 
\begin{equation}
\label{eq:loss}
    \mathcal{L}=\mathcal{L}_{cls}+\beta_1 \mathcal{L}_{smooth}+\beta_2 \mathcal{L}_{con}+\beta_3 \mathcal{L}_{spar}. 
\end{equation}
The overall process of \modelname~is shown in Algorithm~\ref{alg:training}.

\section{Experiment}
In this section, we first evaluate \modelname\footnote{The code of \modelname~is available at \url{https://github.com/RingBDStack/PASTEL}.} on both real-world graphs and synthetic graphs. 
Then we analyze the main mechanisms of \modelname~and the learned structure. 
We mainly focus on the following research questions:
\begin{itemize}[leftmargin=*]
    \item \textbf{RQ1.} 
    How does \modelname~perform in the node classification task? 
    (Section~\ref{sec:classification}) 
    %     How does \modelname~perform in node classification? 
    % (Section~\ref{sec:classification}) 
    \item \textbf{RQ2.} 
    How does the position encoding and the class-wise conflict measure influence the performance of \modelname? 
    (Section~\ref{sec:abla}) 
    \item \textbf{RQ3.} 
    What graph structure \modelname~tend to learn? (Section~\ref{sec:structure}) 
\end{itemize}

\subsection{Experimental Setups}

\begin{table*}[]
\caption{Weighted-F1 score and Macro-F1 score (\% ± standard deviation) of node classification on real-world graph datasets. }
\label{tab:cls}
\resizebox{\linewidth}{!}{%
\centering
\begin{tabular}{cccccccccccccc}
\hline
 &
   &
  \multicolumn{2}{c}{Cora} &
  \multicolumn{2}{c}{Citeseer} &
  \multicolumn{2}{c}{Photo} &
  \multicolumn{2}{c}{Actor} &
  \multicolumn{2}{c}{Chameleon} &
  \multicolumn{2}{c}{Squirrel} \\ \hline
Backbone &
  Model &
  W-F1 &
  M-F1 &
  W-F1 &
  M-F1 &
  W-F1 &
  M-F1 &
  W-F1 &
  M-F1 &
  W-F1 &
  M-F1 &
  W-F1 &
  M-F1 \\ \hline
\multirow{8}{*}{GCN} &
  original &
  79.4±0.9 &
  77.5±1.5 &
  66.3±1.3 &
  62.2±1.2 &
  85.4±2.8 &
  84.6±1.3 &
  21.8±1.3 &
  20.9±1.4 &
  30.5±3.4 &
  30.5±3.3 &
  21.9±1.2 &
  21.9±1.2 \\
 &
  ReNode &
  80.0±0.7 &
  78.4±1.3 &
  66.4±1.0 &
  62.4±1.1 &
  86.2±2.4 &
  85.3±1.6 &
  21.2±1.2 &
  20.2±1.6 &
  30.3±3.2 &
  30.4±2.8 &
  22.4±1.1 &
  22.4±1.1 \\
 &
  AddEdge &
  79.0±0.9&
77.0±1.4&
66.2±1.3&
62.2±1.3&
85.5±1.5&
86.1±1.8&
21.2±1.3&
20.3±1.5&
30.6±1.6&
30.4±1.7&
21.7±1.5&
21.7±1.5 \\
 &
  DropEdge &
  79.8±0.8&
77.8±1.0&
66.6±1.4&
63.4±1.6&
86.8±1.7&
85.4±1.3&
22.4±1.0&
21.4±1.3&
30.6±3.5&
30.6±3.3&
22.8±1.2&
22.8±1.2 \\
 &
  SDRF &
  82.1±0.8 &
  80.6±0.8 &
  69.6±0.4 &
  66.6±0.3 &
  > 5 days &
  > 5 days &
  > 5 days &
  > 5 days &
  39.1±1.2 &
  39.0±1.2 &
  > 5 days &
  > 5 days \\
 &
  NeuralSparse &
  81.7±1.4 &
  80.9±1.4 &
  {\ul 71.8±1.2} &
  {\ul 69.0±1.0} &
  {\ul 89.7±1.9} &
  88.7±1.8 &
  24.4±1.5 &
  {\ul 23.6±1.6} &
  44.9±3.0 &
  44.9±2.8 &
  28.1±1.8 &
  28.1±1.8 \\
 &
  IDGL &
  {\ul 82.3±0.6} &
  {\ul 81.0±0.9} &
  71.7±1.0 &
  68.0±1.3 &
  88.6±2.3 &
  {\ul 88.8±1.4} &
  {\ul 24.9±0.8} &
  22.0±0.7 &
  {\ul 55.4±1.8} &
  {\ul 55.0±1.7} &
  {\ul 28.8±2.3} &
  {\ul 28.9±2.2} \\
 &
  \textbf{\modelname} &
  \textbf{82.5±0.3} &
  \textbf{81.2±0.3} &
  \textbf{72.9±0.8} &
  \textbf{69.3±0.9} &
  \textbf{91.4±2.7} &
  \textbf{91.3±2.2} &
  \textbf{26.4±1.0} &
  \textbf{24.4±1.2} &
  \textbf{57.8±2.4} &
  \textbf{57.3±2.4} &
  \textbf{37.5±0.6} &
  \textbf{37.5±0.7} \\ \hline
\multirow{8}{*}{GAT} &
  original &
  78.3±1.5 &
  76.4±1.7 &
  64.4±1.7 &
  60.6±1.7 &
  88.2±2.9 &
  86.2±2.6 &
  21.8±1.2 &
  20.9±1.1 &
  29.9±3.5 &
  29.9±3.1 &
  20.5±1.4 &
  20.5±1.4 \\
 &
  ReNode &
  78.9±1.2 &
  77.2±1.5 &
  64.9±1.6 &
  61.0±1.5 &
  89.1±2.4 &
  87.1±2.6 &
  21.5±1.2 &
  20.5±1.1 &
  29.2±2.3 &
  29.1±2.0 &
  20.4±1.8 &
  20.4±1.8 \\
 &
  AddEdge &
  78.0±1.6&
76.2±1.6&
64.0±1.3&
60.2±1.3&
88.2±2.4&
86.2±2.5&
21.3±1.2&
20.3±1.1&
29.8±1.7&
29.6±1.5&
20.7±1.6&
20.7±1.6 \\
 &
  DropEdge &
  78.7±1.3&
76.9±1.5&
64.5±1.4&
60.5±1.3&
88.9±1.9&
87.1±2.1&
22.9±1.2&
21.8±1.1&
30.3±1.6&
30.2±1.2&
21.2±1.5&
21.2±1.5 \\
 &
  SDRF &
  77.9±0.7 &
  75.9±0.9 &
  64.9±0.6 &
  {\ul 61.9±0.9} &
  > 5 days &
  > 5 days &
  > 5 days &
  > 5 days &
  43.0±1.9 &
  42.5±1.9 &
  > 5 days &
  > 5 days \\
 &
  NerualSparse &
  {\ul 81.4±4.8} &
  79.4±4.8 &
  64.8±1.5 &
  {\ul 61.9±1.3} &
  {\ul 90.2±2.5} &
  {\ul 88.0±2.3} &
  {\ul 23.4±1.7} &
  \textbf{22.4±1.5} &
  45.6±2.1 &
  45.5±1.8 &
  {\ul 28.8±1.3} &
  {\ul 28.8±1.3} \\
 &
  IDGL &
  80.6±1.0 &
  {\ul 79.7±0.9} &
  {\ul 66.5±1.5} &
  {\ul 61.9±1.9} &
  89.9±3.1 &
  87.7±2.6 &
  22.4±1.5 &
  21.8±1.2 &
  {\ul 48.4±4.0} &
  {\ul 47.8±3.1} &
  27.0±2.6 &
  27.0±2.6 \\
 &
  \textbf{\modelname} &
  \textbf{81.9±1.4} &
  \textbf{80.7±1.2} &
  \textbf{66.6±1.9} &
  \textbf{62.0±1.7} &
  \textbf{91.8±3.2} &
  \textbf{89.4±2.9} &
  \textbf{24.4±2.6} &
  {\ul 22.1±2.6} &
  \textbf{52.1±2.7} &
  \textbf{52.5±2.8} &
  \textbf{35.3±0.9} &
  \textbf{35.3±0.8} \\ \hline
\multirow{8}{*}{APPNP} &
  original &
  80.6±1.6 &
  79.3±1.2 &
  66.5±1.5 &
  62.3±1.5 &
  89.3±1.6 &
  86.3±1.7 &
  21.1±1.5 &
  20.7±1.1 &
  35.3±4.0 &
  35.0±3.8 &
  23.1±1.6 &
  23.1±1.6 \\
 &
  ReNode &
  81.1±0.9 &
  79.9±0.9 &
  66.6±1.7 &
  62.4±1.6 &
  89.6±1.4 &
  87.2±1.3 &
  20.2±2.0 &
  20.0±1.7 &
  33.5±2.5 &
  33.3±2.3 &
  23.9±2.0 &
  23.9±2.0 \\
 &
  AddEdge &
  80.3±1.3&
78.8±1.1&
66.6±2.1&
62.5±2.1&
89.3±1.2&
86.4±1.2&
21.5±1.3&
20.7±1.4&
35.7±1.7&
35.4±1.2&
23.1±1.6&
23.2±1.7 \\
 &
  DropEdge &
  80.9±1.4&
79.4±1.2&
66.7±2.0&
63.0±1.9&
90.0±1.2&
87.0±1.2&
{\ul 21.8±1.8}&
20.8±1.4&
36.0±1.7&
35.7±1.6&
23.3±1.7&
23.3±1.7 \\
 &
  SDRF &
  80.7±0.9 &
  79.1±0.8 &
  {\ul 67.1±0.6} &
  {\ul 63.1±0.8} &
  > 5 days &
  > 5 days &
  > 5 days &
  > 5 days &
  36.5±2.1 &
  35.8±2.1 &
  > 5 days &
  > 5 days \\
 &
  NerualSparse &
  81.1±1.4 &
  79.9±1.2 &
  66.8±1.9 &
  62.7±1.9 &
  91.3±1.8 &
  {\ul 89.4±1.6} &
  {\ul 21.8±1.9} &
  \textbf{21.4±1.5} &
  39.1±2.9 &
  38.7±2.8 &
  28.3±1.5 &
  28.3±1.5 \\
 &
  IDGL &
  {\ul 81.3±0.9} &
  \textbf{80.2±0.9} &
  67.0±1.3 &
  62.9±1.3 &
  {\ul 91.6±1.3} &
  88.6±2.2 &
  21.4±2.4 &
  20.1±2.4 &
  {\ul 41.2±2.2} &
  {\ul 40.6±2.6} &
  {\ul 29.6±2.3} &
  {\ul 29.7±2.2} \\
 &
  \textbf{\modelname} &
  \textbf{82.0±1.0} &
  {\ul 80.0±0.9} &
  \textbf{67.3±1.3} &
  \textbf{63.2±1.5} &
  \textbf{92.3±3.1} &
  \textbf{89.9±2.5} &
  \textbf{22.5±2.0} &
  {\ul 20.9±2.1} &
  \textbf{44.2±3.2} &
  \textbf{43.8±3.4} &
  \textbf{34.6±1.6} &
  \textbf{34.6±1.6} \\ \hline
\multirow{8}{*}{GraphSAGE} &
  original &
  75.4±1.6 &
  74.1±1.6 &
  64.8±1.6 &
  60.7±1.6 &
  86.1±2.5 &
  83.3±2.4 &
  24.0±1.2 &
  23.2±1.0 &
  36.5±1.6 &
  36.2±1.6 &
  27.2±1.7 &
  27.2±1.7 \\
 &
  ReNode &
  76.4±0.9 &
  75.0±1.1 &
  65.4±1.7 &
  61.2±1.7 &
  86.5±1.7 &
  84.1±1.7 &
  23.7±1.2 &
  22.8±1.0 &
  36.4±1.9 &
  36.1±1.9 &
  27.7±1.8 &
  27.7±1.8 \\
 &
  AddEdge &
  75.2±1.2&
73.7±1.2&
65.0±1.4&
60.9±1.3&
86.1±2.8&
83.4±2.6&
23.8±1.7&
23.2±1.6&
36.5±1.5&
36.2±1.3&
26.9±2.1&
26.9±2.1 \\
 &
  DropEdge &
  76.0±1.6&
74.5±1.6&
65.1±1.4&
60.9±1.4&
86.2±1.6&
83.5±1.4&
24.1±1.0&
23.3±0.9&
37.5±1.4&
37.2±1.4&
27.5±1.8&
27.5±1.8 \\
 &
  SDRF &
  75.7±0.8 &
  74.6±0.8 &
  65.3±0.6 &
  \textbf{61.4±0.6} &
  > 5 days &
  > 5 days &
  > 5 days &
  > 5 days &
  41.5±2.6 &
  41.6±2.7 &
  > 5 days &
  > 5 days \\
 &
  NerualSparse &
  {\ul 79.7±1.8} &
  77.8±1.6 &
  64.7±1.4 &
  61.1±1.3 &
  89.1±5.4 &
  {\ul 86.7±5.5} &
  {\ul 25.1±1.2} &
  \textbf{24.4±1.1} &
  39.1±1.9 &
  39.0±1.9 &
  32.2±2.4 &
  32.2±2.4 \\
 &
  IDGL &
  79.2±0.9 &
  {\ul 78.4±0.8} &
  {\ul 65.6±0.9} &
  {\ul 61.3±1.2} &
  {\ul 90.0±1.0} &
  86.3±1.3 &
  24.0±2.6 &
  22.4±2.7 &
  {\ul 43.8±3.4} &
  {\ul 43.0±3.2} &
  {\ul 33.9±0.9} &
  {\ul 33.9±0.8} \\
 &
  \textbf{\modelname} &
  \textbf{81.1±0.8} &
  \textbf{79.8±0.7} &
  \textbf{65.7±1.1} &
  \textbf{61.4±1.4} &
  \textbf{92.0±0.6} &
  \textbf{89.0±1.0} &
  \textbf{26.0±2.4} &
  {\ul 23.6±2.7} &
  \textbf{47.7±0.9} &
  \textbf{46.9±0.9} &
  \textbf{35.5±1.4} &
  \textbf{35.5±1.4} \\ \hline
\end{tabular}
}
\end{table*}

\subsubsection{Datasets}
We conduct experiments on synthetic and real-world datasets to analyze the model's capabilities in terms of both graph theory and real-world scenarios. 
% We conduct experiments on several real-world datasets including both homophily and : 
% Cora, Citeseer and \textit{PubMed}~\cite{sen2008cora} are citation networks of academic papers; 
% \textit{Photo}; 
% \textit{Computers}; 
% \textit{Chameleon};
% \textit{Squirrel};
% \textit{Actor}. 
% \textbf{Real-world Graphs. }
The real-word datasets include various networks with different heterophily degrees to demonstrate the generalization of \modelname. 
% We also conducted experiments on several real-world datasets.  
Cora and Citeseer~\cite{sen2008cora} are citation networks.  
% (1) \textit{Citation network:} Cora, Citeseer and PubMed~\cite{sen2008cora} are citation networks of academic papers, where nodes represent papers, edges denote citation relations, and node labels are academic topics.  
Photo~\cite{shchur2018pitfalls} and and Actor~\cite{pei2020geom} are co-occurrence network. 
Chameleon and Squirrel~\cite{rozemberczki2021multi} are page-page networks in Wikipedia. 
Since we focus on the topology-imbalance issue in this work, we set the number of labeled nodes in each class to be 20.

\subsubsection{Baselines}

We choose representative GNNs as backbones including GCN~\cite{GCN}, GAT~\cite{velivckovic2017graph}, APPNP~\cite{klicpera2018predict}, and GraphSAGE~\cite{hamilton2017inductive}. 
% , ChebNet~\cite{defferrard2016convolutional} and SGC~\cite{wu2019simplifying}. 
The most important baseline is ReNode~\cite{chen2021topology}, which is the only existing work for the topology-imbalance issue. 
We also include some graph structure learning baselines to illustrate the specific effectiveness of \modelname~for the topology-imbalance issue. 
DropEdge~\cite{rong2019dropedge} randomly removes edges at each epoch as structure augmentation. 
To evaluate the effect of increasing the reachability randomly, we use a adding edges method named AddEdge, whose adding strategy is similar to DropEdge. 
SDRF~\cite{topping2021understanding} rewires edges according to their curvatures for the over-squashing issue. 
NeuralSparse~\cite{zheng2020robust} removes potentially task-irrelevant edges for clearer class boundaries. 
IDGL~\cite{chen2020iterative} updates the node representations and structure based on these representations iteratively.

\subsubsection{Parameter Settings}
% For the GNN backbones, we set their depth to be 2 layers and adopt the implementations from the PyTorch Geometric Library\footnote{\url{https://github.com/rusty1s/pytorch_geometric}} in all experiments. 
For the GNN backbones, we set their depth to be 2 layers and adopt the implementations from the PyTorch Geometric Library in all experiments. 
We set the representation dimension of all baselines and \modelname~to be 256. 
We re-implement the NeuralSparse~\cite{zheng2020robust} and SDRF~\cite{topping2021understanding} and the parameters of baseline methods are set as the suggested value in their papers or carefully tuned for fairness. 
For DropEdge and AddEdge, we set the edge dropping/adding probability to 10\%. 
For \modelname, we set the number of heads $m=4$ and the random walk restart probability $\alpha=0.15$. 
The structure fusing coefficients ($\lambda_1$ and $\lambda_2$) and the loss coefficients ($\beta_1$, $\beta_2$ and $\beta_3$) are tuned for each dataset. 
% The start learning rate is 0.01 and the dropout rate is 0.5. 
% All experiments run on the same multi-node GPU cluster, where each node consists of a 64-core Intel Xeon E5-2680v4 CPU @ 2.40GHz with 512GB RAM and a NVIDIA V100 GPU with 32GB RAM. 

\subsection{Evaluation (RQ1)}
\label{sec:classification}

\begin{table}[]
\caption{Weighted-F1 scores and improvements on graphs with different levels of topology-imbalance.}
\label{tab:LMH}
\resizebox{\linewidth}{!}{%
\centering
\begin{tabular}{lcccccc}
\hline
                                     & \multicolumn{2}{c}{Cora-L} & \multicolumn{2}{c}{Cora-M} & \multicolumn{2}{c}{Cora-H} \\ \hline
 &
  \begin{tabular}[c]{@{}c@{}}$RC$\\ 0.4130\end{tabular} &
  \begin{tabular}[c]{@{}c@{}}$SC$\\ -0.6183\end{tabular} &
  \begin{tabular}[c]{@{}c@{}}$RC$\\ 0.4100\end{tabular} &
  \begin{tabular}[c]{@{}c@{}}$SC$\\ -0.6204\end{tabular} &
  \begin{tabular}[c]{@{}c@{}}$RC$\\ 0.4060\end{tabular} &
  \begin{tabular}[c]{@{}c@{}}$SC$\\ -0.6302\end{tabular} \\ \cline{2-7} 
                                     & W-F1 (\%)  & $\Delta$ (\%)  & W-F1 (\%)  & $\Delta$ (\%)  & W-F1 (\%)  & $\Delta$ (\%)  \\ \hline
GCN                       & 80.9±0.9  & —            & 78.8±0.8  & —            & 77.5±1.0  & —            \\
ReNode       & 81.3±0.7  & $\uparrow$0.4         & 79.3±0.8  & $\uparrow$0.5         & 78.3±1.1  & $\uparrow$0.8         \\
SDRF &   81.0±0.7   &   $\uparrow$0.1       &   78.9±0.8    &  $\uparrow$0.1         &    77.9±0.7   &    $\uparrow$0.4      \\
IDGL        & 82.5±1.0  & $\uparrow$1.6         & 80.4±1.0  & $\uparrow$1.6         & 81.6±1.1  & $\uparrow$4.1         \\
\textbf{\modelname}                  & \textbf{82.7±0.9}  & $\uparrow$\textbf{1.8}         & \textbf{81.0±0.9}  & $\uparrow$\textbf{2.2}         & \textbf{81.9±1.1}  & $\uparrow$\textbf{4.4}         \\ \hline
\end{tabular}
}
\end{table}

\subsubsection{\modelname~for Real-world Graphs}
\label{sec:cls_realworld}
We compare \modelname~with the baselines on several datasets on node classification. 
The overall Weighted-F1 (W-F1) scores and the class-balance Macro-F1 (M-F1) scores on different backbones are shown in Table~\ref{tab:cls}. 
The best results are shown in bold and the runner-ups are underlined. 
% From Table~\ref{tab:cls}, we can make the following observations: 
% (1) 
\modelname~shows overwhelming superiority in improving the performance of backbones on all datasets. 
It demonstrates that \modelname~is capable of learning better structures with a more balanced label distribution that reinforces the GNN models. 
% effectively alleviate topology-imbalance and improve GNNs' performance. 
% (2) 
ReNode~\cite{chen2021topology} achieves fewer improvements on datasets of poor connectivity (e.g., CiteSeer) and even damages the performance of backbones on heterophilic datasets (e.g., Chameleon and Actor). 
We think it's because ReNode~\cite{chen2021topology} detects conflicts by Personalized PageRank and fails to reflect the node topological position well when the graph connectivity is poor. 
Besides, ReNode takes the topology boundary as the decision boundary, which is not applicable for heterophilic graphs. 
% (3) 
AddEdge doesn't work in most cases, demonstrating that randomly adding edge is not effective in boosting the reachability. 
The structure augmentation strategy should be carefully designed considering the node relations. 
% (4) 
SDRF~\cite{topping2021understanding} can improve the performance, supporting our intuition that relieving over-squashing helps graph learning. 
But SDRF is still less effective than \modelname~because it only considers the topological properties rather than the supervision information. 
% (5) 
Both NeuralSparse~\cite{zheng2020robust} and IDGL~\cite{chen2020iterative} show good performance among the baselines, showing the effectiveness of learning better structures for downstream tasks. 
However, they are still less effective than \modelname~which 
% considering the supervision information distribution. 
takes the supervision information distribution into consideration. 
\begin{table*}[!tp]
\caption{Weighted-F1 scores (\%) and improvements ($\Delta$) on synthetic SBM graphs with different community structures.}
\label{tab:SBM}
\resizebox{0.92\linewidth}{!}{%
\centering
\begin{tabular}{c|cc|cc|cc|cc|cc|cc|cc}
\hline
 &
  \multicolumn{2}{c|}{SBM-1} &
  \multicolumn{2}{c|}{SBM-2} &
  \multicolumn{2}{c|}{SBM-3} &
  \multicolumn{2}{c|}{SBM-4} &
  \multicolumn{2}{c|}{SBM-5} &
  \multicolumn{2}{c|}{SBM-6} &
  \multicolumn{2}{c}{SBM-7} \\ \hline
$p$ &
  \multicolumn{2}{c|}{0.5000} &
  \multicolumn{2}{c|}{0.5000} &
  \multicolumn{2}{c|}{0.5000} &
  \multicolumn{2}{c|}{0.5000} &
  \multicolumn{2}{c|}{0.5000} &
  \multicolumn{2}{c|}{0.5000} &
  \multicolumn{2}{c}{0.5000} \\
$q$ &
  \multicolumn{2}{c|}{0.0300} &
  \multicolumn{2}{c|}{0.0100} &
  \multicolumn{2}{c|}{0.0083} &
  \multicolumn{2}{c|}{0.0071} &
  \multicolumn{2}{c|}{0.0063} &
  \multicolumn{2}{c|}{0.0056} &
  \multicolumn{2}{c}{0.0050} \\ \hline
$RC$ &
  \multicolumn{2}{c|}{0.4979} &
  \multicolumn{2}{c|}{0.4984} &
  \multicolumn{2}{c|}{0.4990} &
  \multicolumn{2}{c|}{0.4994} &
  \multicolumn{2}{c|}{0.5002} &
  \multicolumn{2}{c|}{0.5004} &
  \multicolumn{2}{c}{0.5009} \\
$SC$ &
  \multicolumn{2}{c|}{0.0998} &
  \multicolumn{2}{c|}{0.0999} &
  \multicolumn{2}{c|}{0.1000} &
  \multicolumn{2}{c|}{0.1001} &
  \multicolumn{2}{c|}{0.1007} &
  \multicolumn{2}{c|}{0.1017} &
  \multicolumn{2}{c}{0.1144} \\ \hline
 &
  W-F1 &
  $\Delta$ &
  W-F1 &
  $\Delta$ &
  W-F1 &
  $\Delta$ &
  W-F1 &
  $\Delta$ &
  W-F1 &
  $\Delta$ &
  W-F1 &
  $\Delta$ &
  W-F1 &
  $\Delta$ \\ \hline
GCN &
  40.29 &
  — &
  42.37 &
  — &
  42.99 &
  — &
  44.13 &
  — &
  45.19 &
  — &
  45.21 &
  — &
  45.22 &
  — \\
ReNode &
  41.33 &
  $\uparrow$1.04 &
  42.40 &
  $\uparrow$0.03 &
  43.21 &
  $\uparrow$0.22 &
  44.56 &
  $\uparrow$0.43 &
  45.20 &
  $\uparrow$0.01 &
  45.08 &
  $\downarrow$0.13 &
  44.89 &
  $\downarrow$0.33 \\
\textbf{\modelname} &
  \textbf{45.67} &
  \textbf{$\uparrow$5.38} &
  \textbf{57.61} &
  \textbf{$\uparrow$15.24} &
  \textbf{58.33} &
  \textbf{$\uparrow$15.34} &
  \textbf{60.29} &
  \textbf{$\uparrow$16.16} &
  \textbf{66.41} &
  \textbf{$\uparrow$21.22} &
  \textbf{66.45} &
  \textbf{$\uparrow$21.24} &
  \textbf{66.57} &
  \textbf{$\uparrow$21.35} \\ \hline
\end{tabular}
}
\end{table*}
\subsubsection{\modelname~under Different Levels of Topology-imbalance}

To further analyze \modelname's ability in alleviating the topology-imbalance issue, we verify the \modelname~under different levels of topology-imbalance. 
% Since motivation of relieving topology-imbalance, 
We randomly sampled 1,000 training sets and calculate the reaching coefficient $RC$ and squashing coefficient $SC$ as introduced in Section~\ref{sec:understand}. 
Then we choose 3 training sets with different levels of topology-imbalance according to the conclusion in Section~\ref{sec:quantitative} and we denote them as Cora-L, Cora-M, and Cora-H, according to the degree of topology imbalance. 
Note that larger $RC$ means better reachability and larger $SC$ means lower squashing. 
We evaluate \modelname~and several baselines with the GCN as the backbone and show the dataset information, the Weighted-F1 scores, and their improvements ($\Delta$) over the backbones in Table~\ref{tab:LMH}. 
% We can make the following observations: 
% (1) 
The performance of node representation learning generally gets worse with the increase of the topology-imbalance degree of the dataset. 
% (2) 
Both the node re-weighting method (i.e., ReNode~\cite{chen2021topology}) and the structure learning methods (i.e., IDGL~\cite{chen2020iterative}, SDRF~\cite{topping2021understanding} and \modelname) can achieve more improvement with the increase of dataset topology-imbalance. 
% (3) 
\modelname~performs best on all datasets with different degrees of topology-imbalance and it can achieve up to 4.4\% improvement on the highly topology-imbalance dataset.

\subsubsection{\modelname~for Synthetic Graphs}
\begin{figure}[t]
    \centering
    \includegraphics[width=0.95\linewidth]{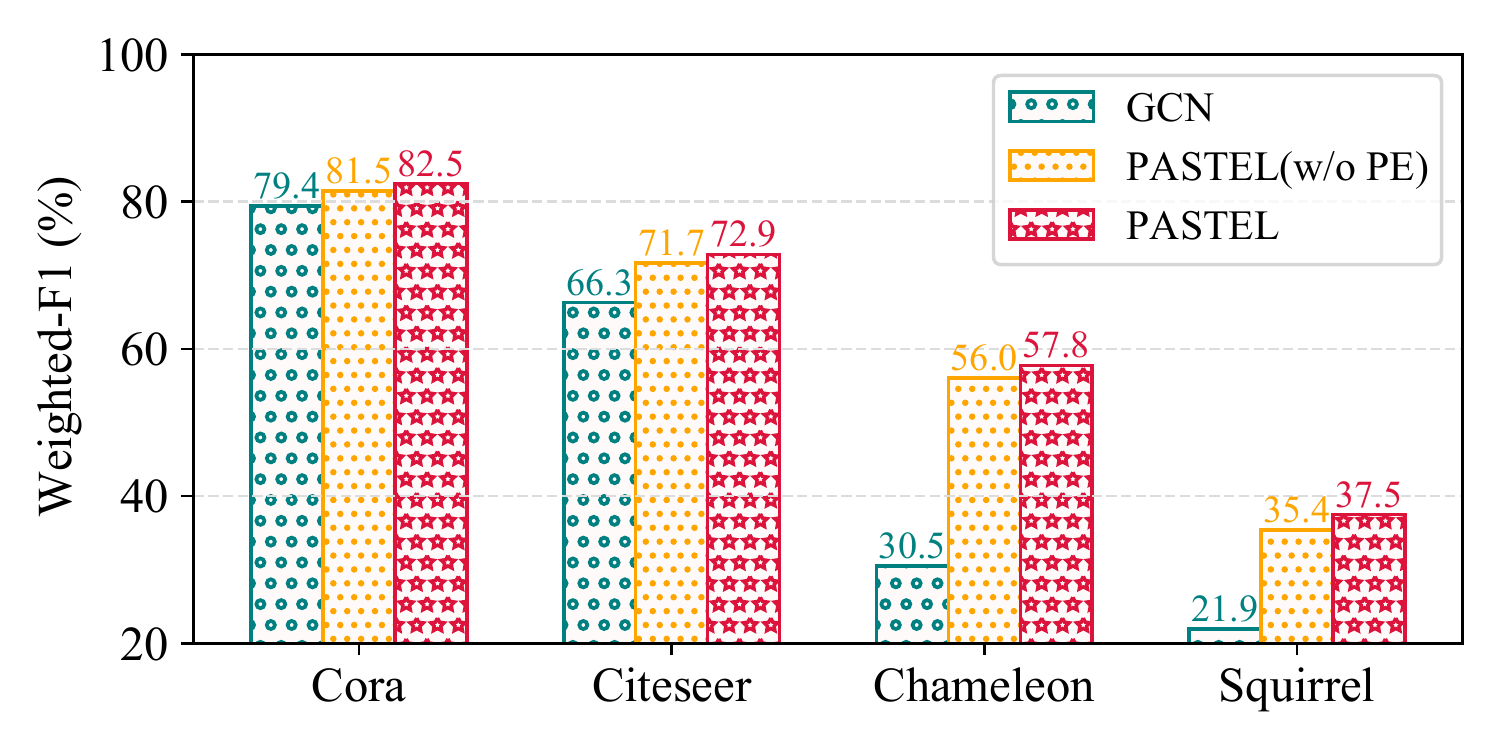}
    \caption{The impact of position encoding. }
    \label{fig:abla_position}
\end{figure}
We generate 7 synthetic graph datasets with different community structures using the Stochastic Block Model (SBM) $\mathcal{G}(N,C,p,q)$~\cite{holland1983stochastic}, where the number of nodes $N=3000$, the number of community $C=6$, $p$ denotes the edge probability within a community and $q$ denotes the edge probability between communities. 
We show the classification Weighted-F1 scores and improvements are shown in Table~\ref{tab:SBM}. 
% We can have the following observations:
% (1) 
With a more clear community structure, the reaching coefficient $RC$ increases and the squashing coefficient $SC$ also increases, leading to the increase of GCN's performance, which agrees with the conclusion obtained in Section~\ref{sec:quantitative}. 
% (2) 
ReNode shows unsatisfied performance in boosting the node classification. 
% (3) 
\modelname~can increase the classification weighted-F1 score by 5.38\%-21.35\% on SBM graphs with different community structures, showing superior effectiveness. 

% \subsubsection{\modelname~for Large-scale Graphs}
% We conduct experiments on a large-scale dataset \textit{ogbn-arxiv}~\cite{HuFZDRLCL20}. 

% PPRGo~\cite{bojchevski2020scaling}
% achor-GCN~\cite{chen2020iterative}

% \subsection{Ablation Study}
\subsection{Analysis of \modelname~(RQ2)}
\label{sec:abla}
We conduct ablation studies for the two main mechanisms of \modelname, position encoding and class-wise conflict measure. 
% , to provide further analysis for \modelname. 
\subsubsection{Impact of the Position Encoding}
\label{sec:abla_position}

We design an anchor-based position encoding mechanism in Section~\ref{sec:position_encoding}, which reflects the relative topological position to labeled nodes and further maximizes the label influence within a class. 
To evaluate the effectiveness of position encoding, we compare \modelname~with a variant \textbf{\modelname~(w/o PE)}, which removes the position encoding and directly take the node features for metric learning in Eq.~\eqref{eq:metric}. 
% \begin{itemize}[leftmargin=*]
%     \item 
%     \textbf{\modelname~(w/o PE)}, which removes the position encoding and directly take the node features for metric learning in Eq.~\eqref{eq:metric}. 
% \end{itemize}
\begin{figure}[!tp]
    \centering
    \includegraphics[width=0.95\linewidth]{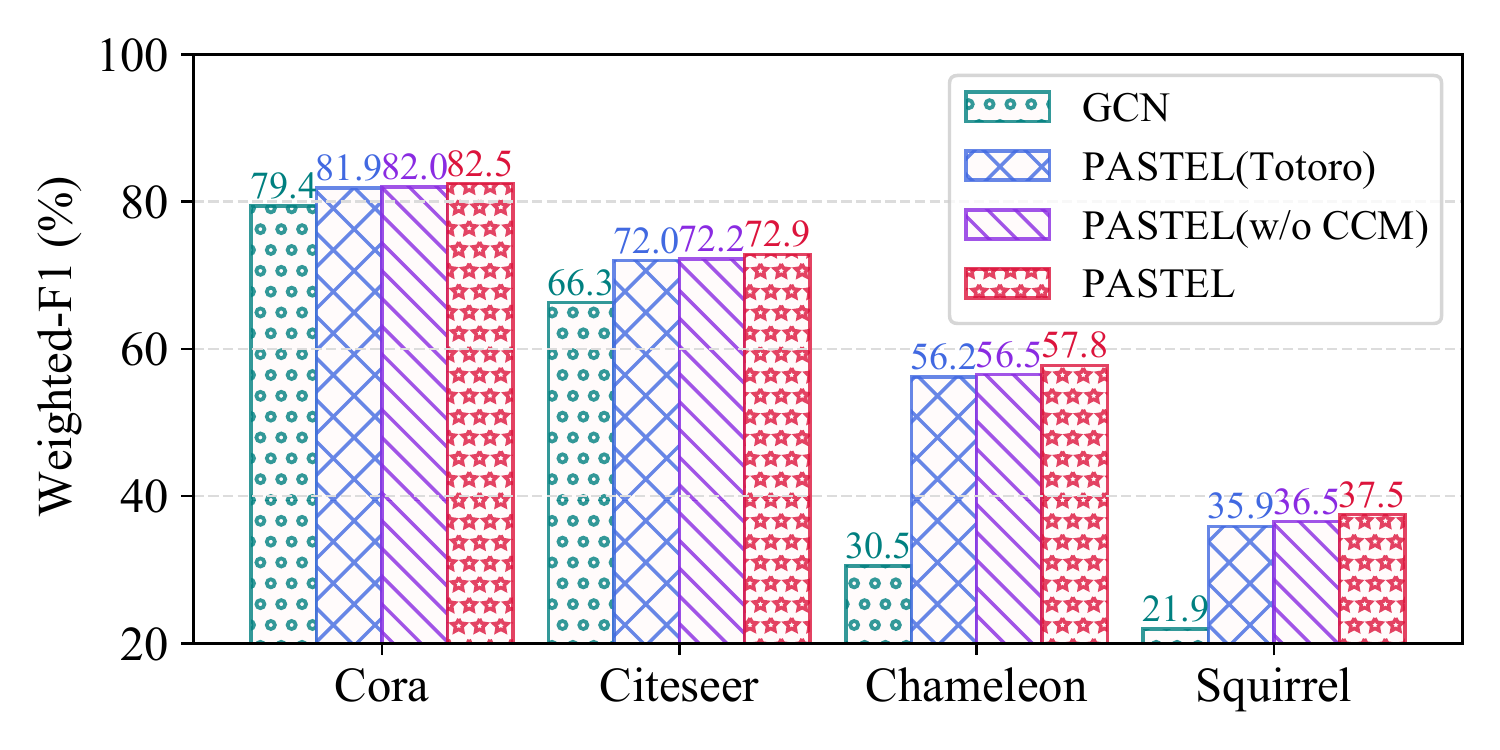}
    \caption{The impact of class-wise conflict measure. }
    \label{fig:abla_conflict}
\end{figure}
Here we use the GCN as the backbone. 
% The comparison results are shown in Figure~\ref{fig:abla_position}. 
% As we can observe, 
As shown in Figure~\ref{fig:abla_position}, the structure learning strategy of \modelname~contributes the most, which can achieve at most 25.5\% improvement in terms of Weighted-F1 score with only node features. 
Although \modelname~(w/o PE) effectively improves the performance of backbones to some extent, the position encoding still benefits learning better structure to relieve the topology-imbalance with 1.0\%-1.8\% improvements than \modelname~(w/o PE). 

\begin{figure*}[!htp]
\centering
\subfigure[{Original Graph. }]{
\begin{minipage}[t]{ 0.1875\linewidth}
\centering
\includegraphics[width=\linewidth]{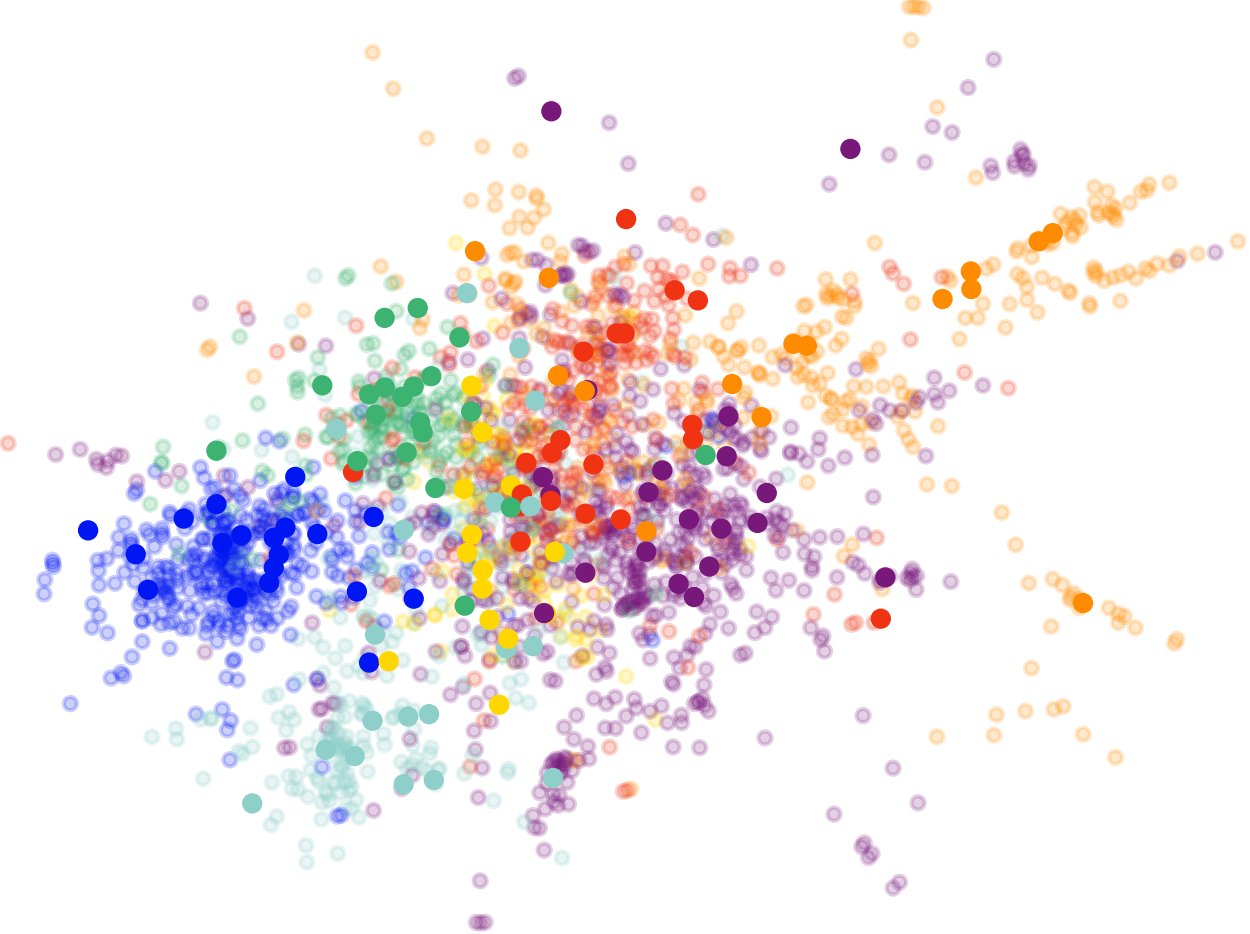}
\label{fig:cora}
\end{minipage}
}
\subfigure[ReNode. ]{
\begin{minipage}[t]{ 0.1875\linewidth}
\centering
\includegraphics[width=\linewidth]{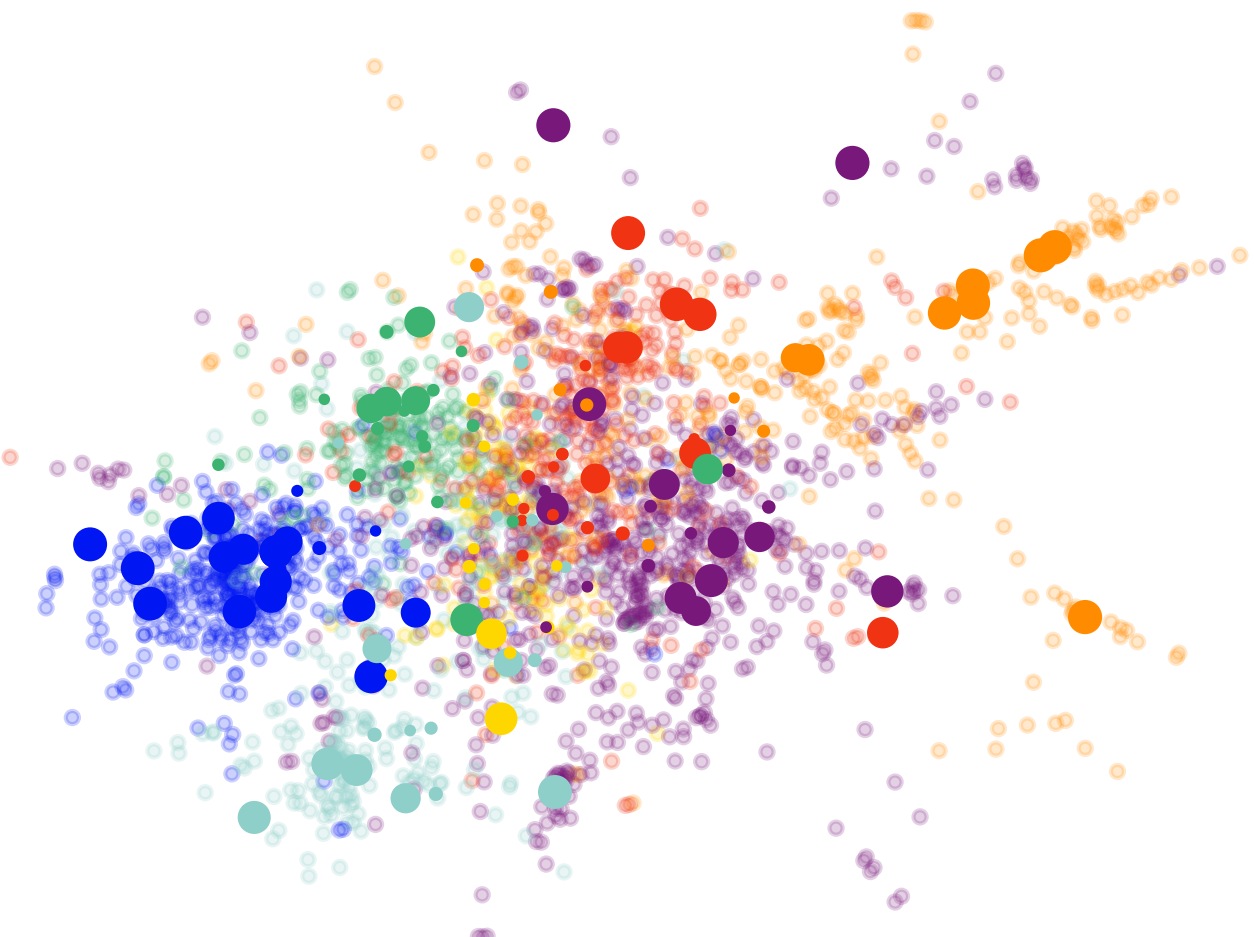}
\label{fig:renode}
\end{minipage}
}
\subfigure[SDRF. ]{
\begin{minipage}[t]{ 0.1875\linewidth}
\centering
\includegraphics[width=\linewidth]{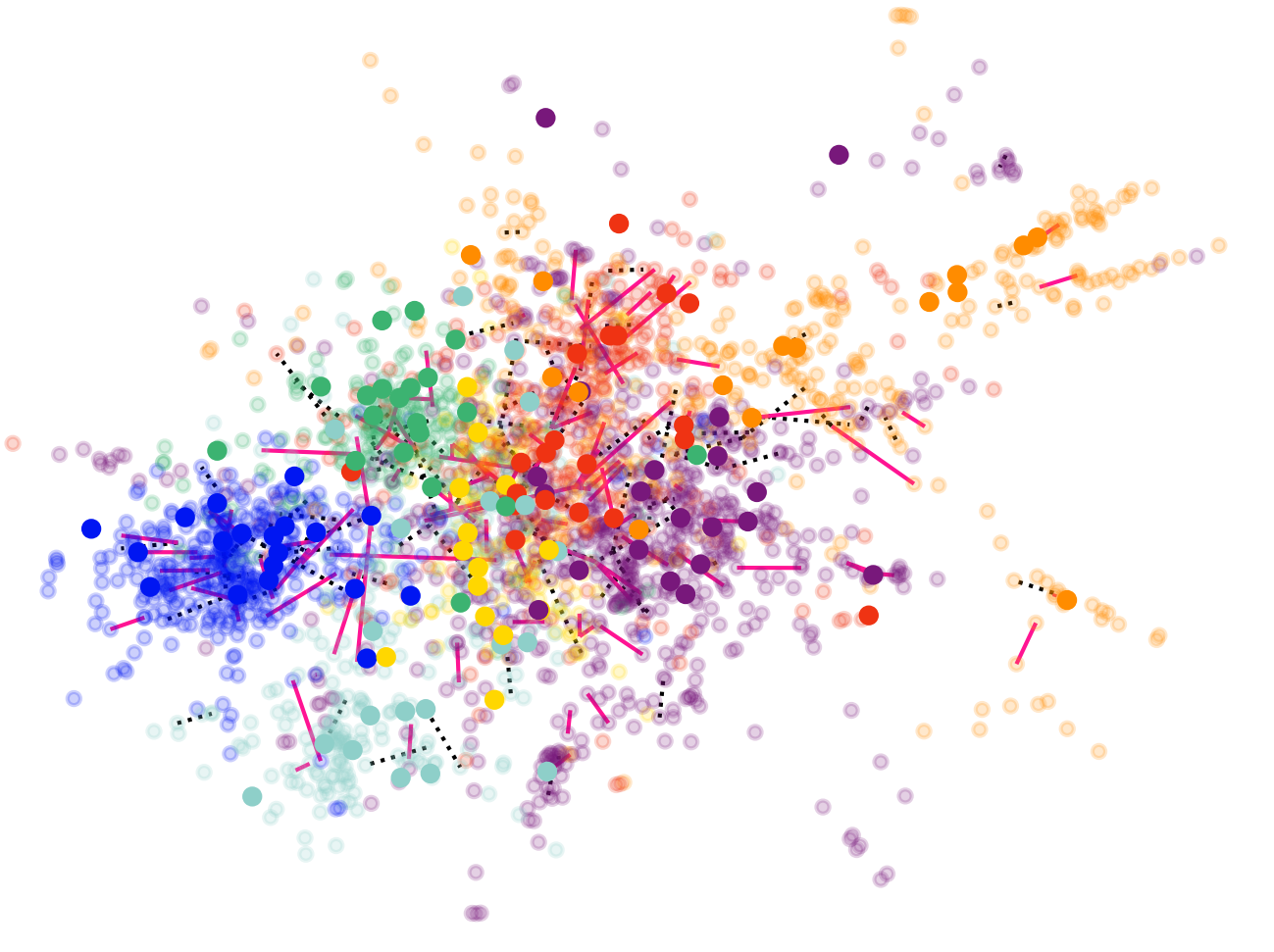}
\label{fig:SDRF}
\end{minipage}
}
\subfigure[IDGL. ]{
\begin{minipage}[t]{ 0.1875\linewidth}
\centering
\includegraphics[width=\linewidth]{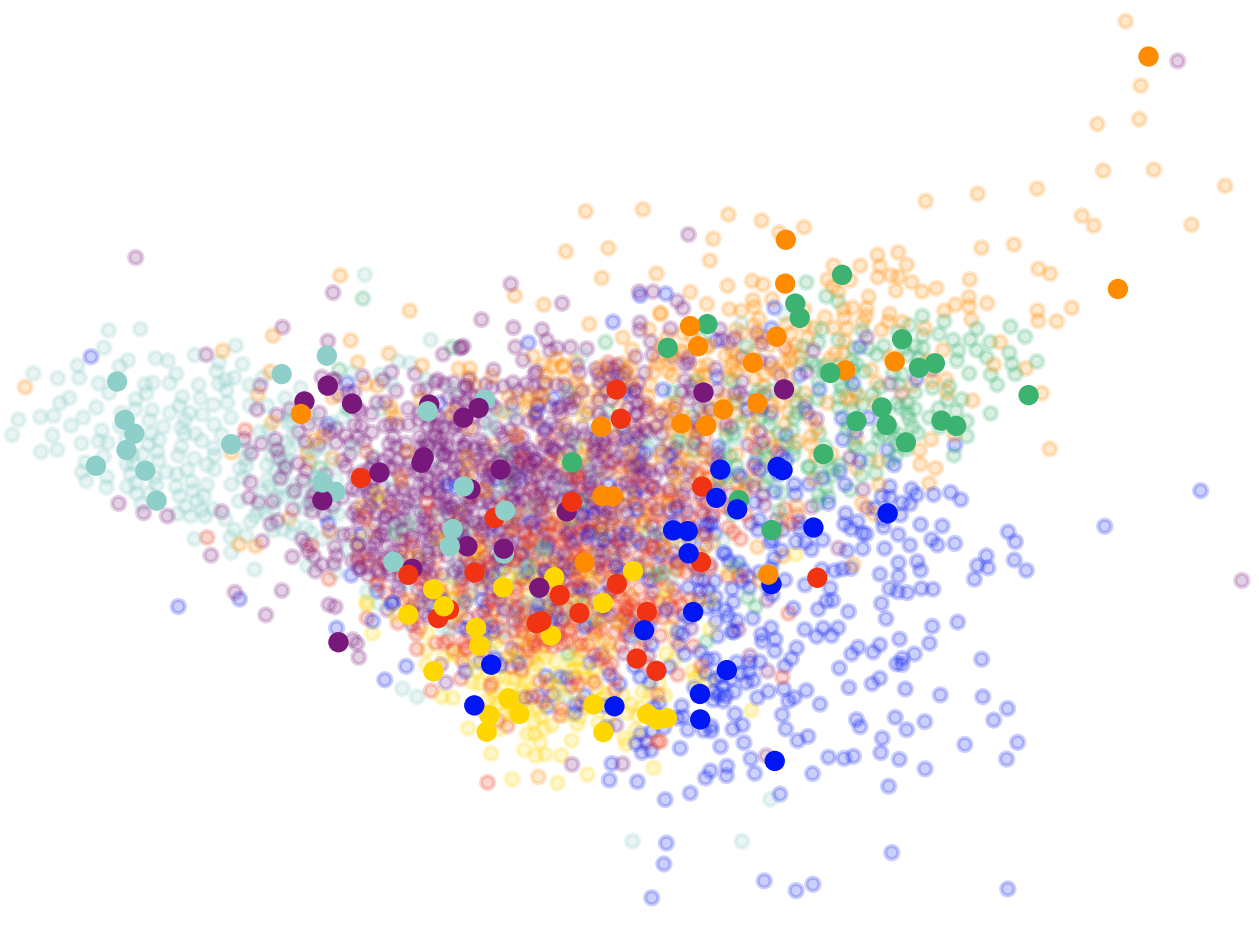}
\label{fig:IDGL}
\end{minipage}
}
\subfigure[\modelname. ]{
\begin{minipage}[t]{ 0.1875\linewidth}
\centering
\includegraphics[width=\linewidth]{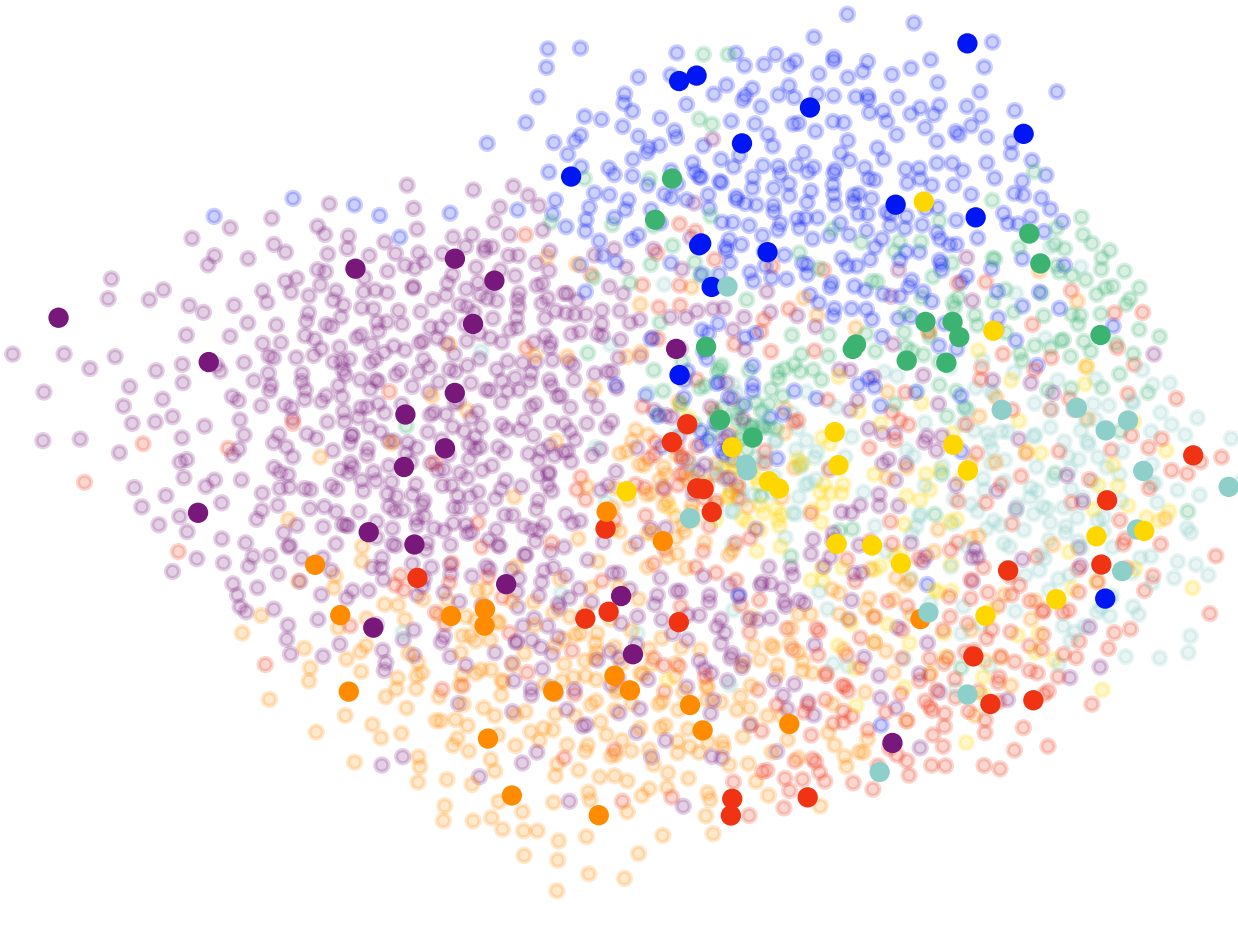}
\label{fig:ours}
\end{minipage}
}
\centering
\caption{Structure visualization. (a) Original graph of Cora and learned graphs by (b) ReNode, (c) SDRF, (d) IDGL and (e) \modelname. }
\label{fig:visual}
\end{figure*}

\subsubsection{Impact of the Class-wise Conflict Measure}
\label{sec:abla_conflict}

We designed a class-wise conflict measure in Section~\ref{sec:conflict} as edge weights to guide learning structures with better intra-class connectivity. 
Here, we compare \modelname~with its two variants to analyze the impact of the class-wise conflict measure: 
(1) \textbf{\modelname~(w/o CCM)}, which removes the class-wise conflict measure and directly takes the learned edge possibilities in Eq.~\eqref{eq:metric} as the edge weights. 
(2) \textbf{\modelname~(Totoro)}, which takes the Totoro metric introduced in ReNode~\cite{chen2021topology} as the conflict measure of nodes in Eq.~\eqref{eq:conflict_weight}. 
% \begin{itemize}[leftmargin=*]
%     \item 
%     \textbf{\modelname~(w/o CCM)}, which removes the class-wise conflict measure and directly takes the learned edge possibilities in Eq.~\eqref{eq:metric} as the edge weights. 
%     \item 
%     \textbf{\modelname~(Totoro)}, which takes the Totoro metric introduced in ReNode~\cite{chen2021topology} as the conflict measure of nodes in Eq.~\eqref{eq:conflict_weight}. 
% \end{itemize}
Here we use the GCN as the backbone. 
The comparison results are shown in Figure~\ref{fig:abla_conflict}. 
On four datasets, \modelname~consistently outperforms the other two variants. 
Even without the conflict measure, \modelname~(w/o CCM) still shows better performance than \modelname~(Totoro), indicating the limitation of ReNode when capturing the relative topology positions without clear homophily structures. 

% \begin{figure}
%     \centering
%     \includegraphics[width=\linewidth]{figs/GPR.pdf}
%     \caption{The change of Group PageRank value. }
%     \label{fig:gpr}
% \end{figure}

% \subsection{Parameter Sensitivity}

\subsection{Analysis of Learned Structure (RQ3)}
\label{sec:structure}
We analyze the learned graph by \modelname~in terms of visualization and structural properties.  
% to provide further insights into the changes brought by \modelname. 
% ~more intuitively. 

\subsubsection{Structure Visualization}
% We visualize the learned graph structure to provide further insights into the changes brought by \modelname~more intuitively. 
In Figure~\ref{fig:visual}, we visualize the original graph of Cora and the graphs learned by ReNode~\cite{chen2021topology}, SDRF~\cite{topping2021understanding}, IDGL~\cite{chen2020iterative} and \modelname~using \textit{networkx}. 
For clarity, the edges are not shown. 
The solid points denote the labeled nodes, the hollow points denote the unlabeled nodes, and the layout of nodes denotes their connectivities. 
The node size in Figure~\ref{fig:renode} denotes the learned node weight in ReNode, and the solid lines and dashed lines in Figure~\ref{fig:SDRF} denote the added and deleted edges by SDRF, respectively. 
As we can observe, ReNode gives more weights to nodes in the topology center of each class and SDRF tends to build connections between distant or isolated nodes. 
% Both IDGL and \modelname~iteratively update the graph structure and node representations. 
Even though the structure learned by IDGL can make the nodes of a class close, there are still some overlapping and entangled areas between classes. 
Benefiting from the position encoding and class-wise conflict measure, \modelname~can obtain graph structure with clearer class boundaries. 

\begin{table}[tp]
\caption{Properties and performance of the original graph and learned graphs of Cora.}
\label{tab:graph_property}
\resizebox{\linewidth}{!}{%
\centering
\begin{tabular}{cccccc}
\hline
          & Original Graph & ReNode & SDRF  & IDGL  & \textbf{\modelname} \\ \hline
$RC$      & 0.4022          & 0.4022  & 0.4686 & 0.5028 & 0.5475               \\
$SC$      & -0.6299          & -0.6299  & -0.4942 & -0.4069 & -0.3389               \\ \hline
W-F1 (\%) & 79.44          & 80.34  &  82.01 & 82.38 & \textbf{82.86}               \\ \hline
\end{tabular}
}
\end{table}
\subsubsection{Change of $RC$ and $SC$}
We also show the reaching coefficient $RC$ and the squashing coefficient $SC$ of the above graphs in Figure~\ref{fig:visual} and the Weighted-F1 score learned on them in Table~\ref{tab:graph_property}. 
Here we choose the GCN as the model backbone. 
All of the structure learning methods (SDRF~\cite{topping2021understanding}, IDGL~\cite{chen2020iterative} and \modelname) learn structures with larger reaching coefficient and larger squashing coefficient, leading the performance improvement of node classification. 
This phenomenon supports our propositions in Section~\ref{sec:quantitative} again. 

\subsubsection{Change of GPR Vector}
\label{sec:gpr_visual}
\begin{figure}[tp]
\centering
\subfigure[Original Graph.]{
\begin{minipage}[t]{ 0.43\linewidth}
\centering
\includegraphics[width=\linewidth]{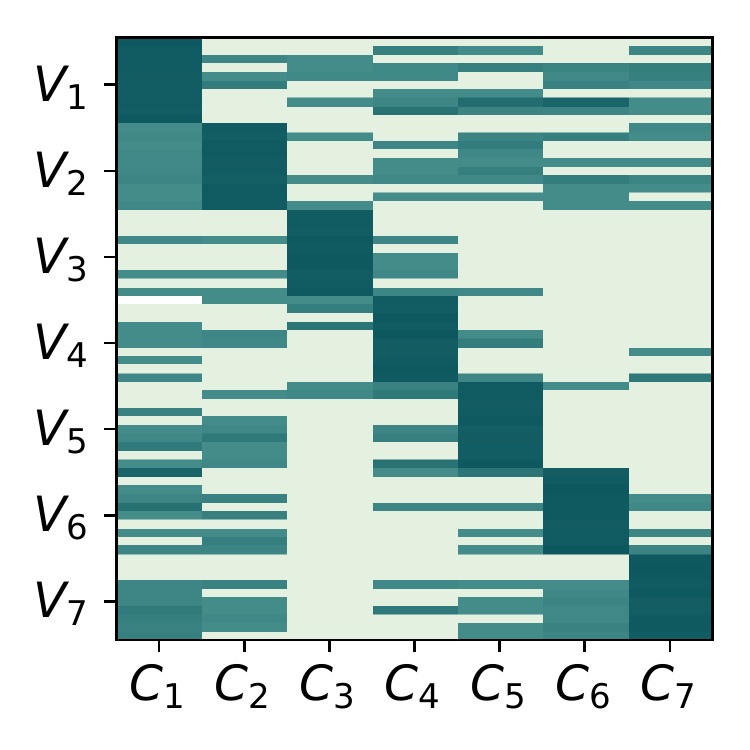}
\label{fig:gpr1}
%\caption{fig2}
\end{minipage}
}
\subfigure[Learned Graph.]{
\begin{minipage}[t]{ 0.43\linewidth}
\centering
\includegraphics[width=\linewidth]{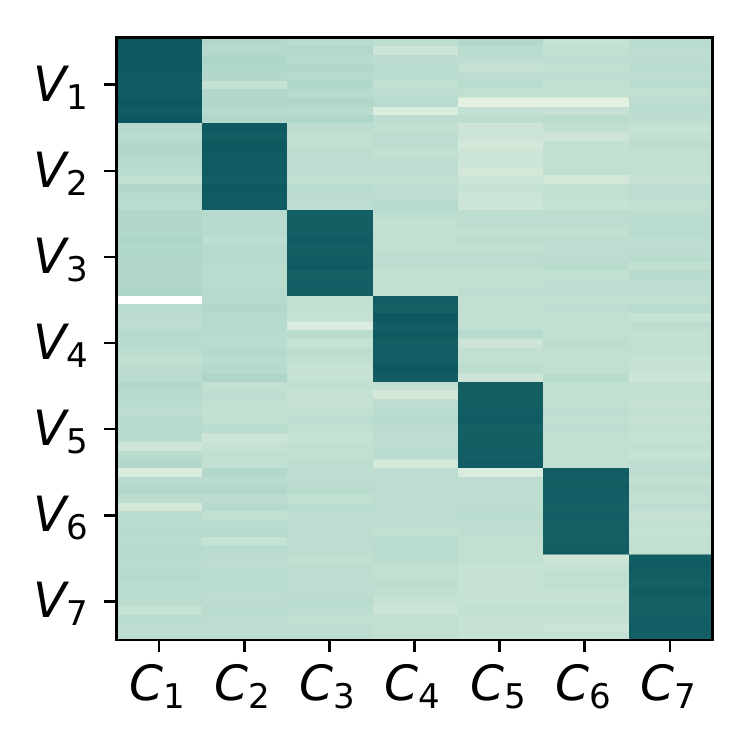}
\label{fig:gpr2}
\end{minipage}
}
\centering
\caption{Heat maps for the Group PageRank value of (a) the original graph and (b) the learned graph by \modelname. }
\label{fig:gpr}
\end{figure}

The class-wise conflict measure is calculated by the Group PageRank (GPR), which reflects the label influence of each class. 
% We show the change of Group PageRank vectors brought by \modelname~in Figure~\ref{fig:gpr}. 
We randomly choose 10 nodes for each class in Cora and show their GPR vectors $\mathbf{P}^{gpr}_{i}$ in the original graph in Figure~\ref{fig:gpr1} and the learned graph in Figure~\ref{fig:gpr2}, respectively, where the color shade denotes the magnitude, $V_i$ denotes 10 nodes of class $i$ and $C_i$ denotes the $i$-th class. 
% $\mathbf{P}^{gpr}_{i}\in\mathbb{R}^{1\times C}$
In Figure~\ref{fig:gpr1}, the off-diagonal color blocks are also dark, indicating that the label influence of each class that nodes obtained from the original graph is still entangled to some extent, which could bring difficulties to the GNN optimization. 
After the structure learning guided by the proposed class-wise conflict measure, Figure~\ref{fig:gpr2} exhibits 7 clear diagonal blocks and the gaps between the diagonal and off-diagonal block are widened, indicating that nodes can receive more supervision information of its ground-truth class. 
We can further make a conclusion that the class-wise conflict measure plays an important role on giving guidance for more class connectivity orthogonality.

\section{Conclusion}
We proposed a novel framework named \modelname~for the graph topology-imbalance issue. 
We provide a new understanding and two quantitative analysis metrics of topology-imbalance in the perspective of under-reaching and over-squashing, answering the questions that how topology-imbalance affects GNN's performance as well as what graphs are susceptible to it. 
\modelname~designs an anchor-based position encoding mechanism and a class-wise conflict measure to obtain structures with better in-class connectivity. 
% , thereby relieving the under-reaching and over-squashing phenomena. 
Comprehensive experiments demonstrate the potential and adaptability of \modelname. 
An interesting future direction is to incorporate the proposed two quantitative metrics into the learning process to address topology-imbalance more directly. 

\begin{acks}
The corresponding author is Jianxin Li. 
The authors of this paper were supported by the NSFC through grant 6187202 and the Academic Excellence Foundation of BUAA for PhD Students. 
Philip S. Yu was supported by NSF under grants III-1763325, III-1909323, III-2106758, and SaTC-1930941. 

% U20B2053
\end{acks}
% \bibliographystyle{ACM-Reference-Format}

\clearpage
\bibliographystyle{ACM-Reference-Format}
\balance
\bibliography{ref}

\end{document}